\documentclass[lettersize,journal]{IEEEtran}
\usepackage{amsmath,amsfonts}
\usepackage{algorithmic}
\usepackage{algorithm}
\usepackage{array}
\usepackage[caption=false,font=normalsize,labelfont=sf,textfont=sf]{subfig}
\usepackage{textcomp}
\usepackage{stfloats}
\usepackage{url}
\usepackage{verbatim}
\usepackage{graphicx}
\usepackage{cite}
\usepackage{tikz}
\usepackage{comment}
\usepackage{amsmath}
\usepackage{amsfonts}
\usepackage{bbding}
\usepackage{pgfplots}
\usepackage{amssymb}
\usepackage{booktabs}
\usepackage{diagbox}
\usepackage{makecell}
\usepackage{multirow}
\usepackage{hyperref}
\usepackage{xcolor}  
\usepackage{colortbl}  
\newcommand{\etal}{\textit{et al.}}

\begin{document}

\title{DirectMHP: Direct 2D Multi-Person Head Pose Estimation with Full-range Angles}

\author{Huayi Zhou, Fei Jiang, and Hongtao Lu, \IEEEmembership{Member, IEEE}
\thanks{H. Zhou, and H. Lu are with Department of Computer Science and Engineering, Shanghai Jiao Tong University, Shanghai 200240, China (e-mail: sjtu\_zhy@sjtu.edu.cn;htlu@sjtu.edu.cn).}
\thanks{F. Jiang is with Intelligent of AI Education, East China Normal University, Shanghai 200062, China (e-mail: fjiang@mail.ecnu.edu.cn).}
\thanks{Manuscript received February 14, 2023; revised May 25, 2023.}}

\markboth{Journal of \LaTeX\ Class Files,~Vol.~14, No.~8, April~2023}%
{Zhou \MakeLowercase{\textit{et al.}}: A Sample Article Using IEEEtran.cls for IEEE Journals}

\IEEEpubid{~\copyright~2023 IEEE \qquad\qquad\qquad\quad}

\maketitle

\begin{abstract}
Existing head pose estimation (HPE) mainly focuses on single person with pre-detected frontal heads, which limits their applications in real complex scenarios with multi-persons. We argue that these single HPE methods are fragile and inefficient for Multi-Person Head Pose Estimation (MPHPE) since they rely on the separately trained face detector that cannot generalize well to full viewpoints, especially for heads with invisible face areas. In this paper, we focus on the full-range MPHPE problem, and propose a direct end-to-end simple baseline named DirectMHP. Due to the lack of datasets applicable to the full-range MPHPE, we firstly construct two benchmarks by extracting ground-truth labels for head detection and head orientation from public datasets AGORA and CMU Panoptic. They are rather challenging for having many truncated, occluded, tiny and unevenly illuminated human heads. Then, we design a novel end-to-end trainable one-stage network architecture by joint regressing locations and orientations of multi-head to address the MPHPE problem. Specifically, we regard pose as an auxiliary attribute of the head, and append it after the traditional object prediction. Arbitrary pose representation such as Euler angles is acceptable by this flexible design. Then, we jointly optimize these two tasks by sharing features and utilizing appropriate multiple losses. In this way, our method can implicitly benefit from more surroundings to improve HPE accuracy while maintaining head detection performance. We present comprehensive comparisons with state-of-the-art single HPE methods on public benchmarks, as well as superior baseline results on our constructed MPHPE datasets. Datasets and code are released in \url{https://github.com/hnuzhy/DirectMHP}.
\end{abstract}



\section{Introduction}\label{intro}

The task of head pose estimation (HPE) is to estimate the orientation (e.g., Euler angles) of human heads from images or videos. It has numerous applications, such as identifying customer behavior \cite{de2013robust}, monitoring driver behavior \cite{murphy2010head}, assisting classroom observation \cite{ahuja2019edusense}, and treating as a proxy of eye-gaze/attention \cite{ahuja2021classroom, langton2004influence}. Although HPE has been extensively studied for a long time \cite{murphy2008head, wu2019facial}, most current methods focus on head pose of the frontal or large-angle face \cite{ruiz2018fine, yang2019fsa, valle2020multi, hsu2018quatnet, mo2021osgg, zhang2020fdn, behera2020rotation}, instead of heads with invisible face parts. This is because heads with visible face areas naturally have richer visual features. Meanwhile, visible face is also an essential factor for face detection and face alignment when implementing most HPE approaches with two separate stages (face detection/alignment, and orientation estimation) in real applications.
 
\begin{figure}[!t]
\centering
\includegraphics[width=\columnwidth]{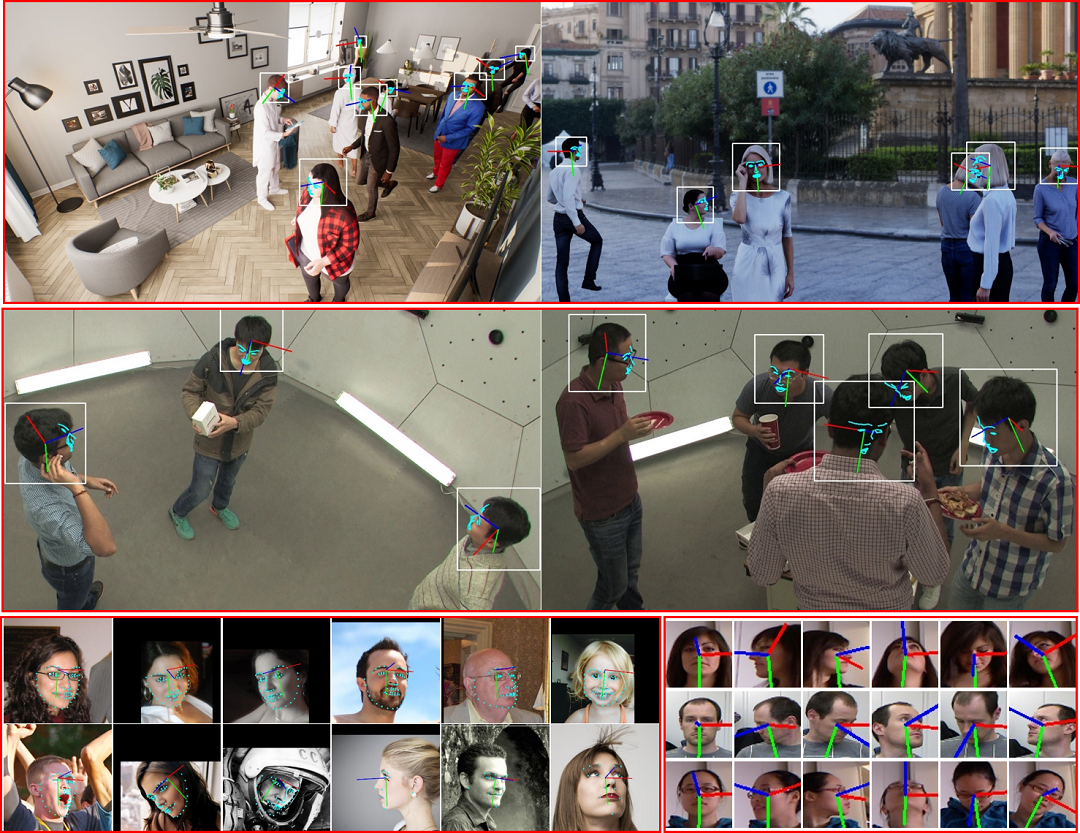}
\caption{Example images of two constructed challenging datasets: AGORA-HPE ({\it top row}) and CMU-HPE ({\it middle row}), and two widely used single HPE datasets: 300W-LP \& AFLW2000 ({\it left-down}) and BIWI ({\it right-down}). Head pose label is plotted by pitch (red-axis), yaw (green-axis) and roll (blue-axis) angles in indicated directions. Face landmarks (if given) are drawn in cyan. Please zoom in for best views.}
\label{datasetexamples}
\end{figure}

Up to now, the two most widely used HPE datasets BIWI \cite{fanelli2013random} and 300W-LP \cite{zhu2017face} contain head pose labels only in {\it narrow-range} yaw angles $(-99^{\circ}, 99^{\circ})$. Examples are shown in Fig.~\ref{datasetexamples}. The performance of existing HPE methods tends to be {\it saturated} on these benchmarks. However, HPE methods covering {\it full-range} yaw angles $(-180^{\circ}, 180^{\circ})$ thus including back-head equally have vital application value. For example, student faces may be invisible in classroom attention observation, and arbitrary head orientation of visitors should be captured in unmanned retail. WHENet \cite{zhou2020whenet} firstly points out this deficiency and obtains an augmented full-range HPE benchmark by combining 300W-LP \cite{zhu2017face} with the CMU Panoptic Dataset \cite{joo2015panoptic}. Then, it proposes a modified HopeNet \cite{ruiz2018fine} to tackle the full-range single HPE problem.

\begin{figure}[!t]
\centering
\includegraphics[width=\columnwidth]{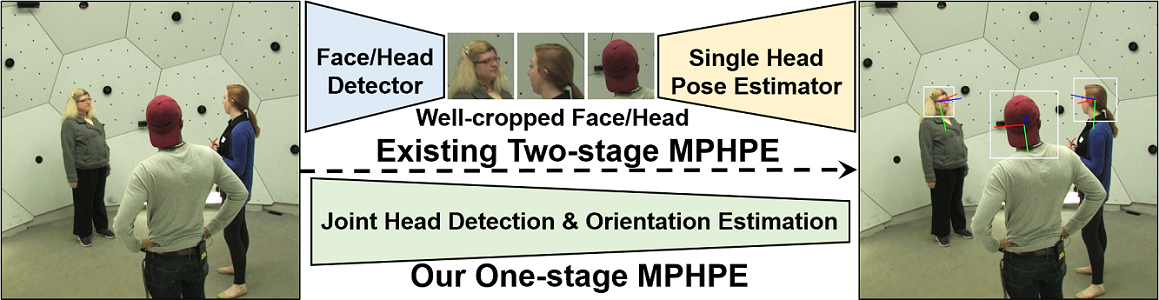}
\caption{The pipeline illustration of existing HPE methods, which are all based on two separate stages, and our proposed one-stage MPHPE.}
\label{illustration}
\end{figure}

Nevertheless, we claim that the current two-stage HPE methods (see Fig.~\ref{illustration}) have the following two shortcomings: (1) Two-stage models of face/head detection and orientation estimation cannot be trained end-to-end, which makes it incompact and inefficient; (2) Separate stages cannot integrate and exploit information of the whole human body and the surrounding background, thus these models are not robust to challenging conditions when applied in-the-wild. Unfortunately, to the best of our knowledge, there are no dedicated public datasets for the Multi-Person Head Pose Estimation (MPHPE) task. Likewise, no public methods try to tackle the MPHPE problem by the direct end-to-end training way.

Facing these challenges, based on two public human-related datasets AGORA \cite{patel2021agora} and CMU Panoptic \cite{joo2015panoptic} with provided 3D face landmarks and camera parameters, we firstly construct two 2D full-range MPHPE datasets AGORA-HPE and CMU-HPE, respectively. Then, we propose a novel one-stage end-to-end network structure DirectMHP which can directly predict the full-range poses of all human heads appearing in one image. The contributions of our work are:

\begin{itemize}
\item We bring forward the 2D full-range Multi-Person Head Pose Estimation (MPHPE) problem for the first time.
\item We construct two automatic labeled reasonable MPHPE benchmarks based on two corresponding human-related datasets AGORA and CMU Panoptic.
\item We propose a novel end-to-end trainable method named DirectMHP for the MPHPE, which achieves joint head detection and pose estimation by sharing features.
\item We integrate head pose as an adjacent attribute into the typical object prediction, which supports arbitrary pose representation such as Euler angles. 
\item We qualitatively and quantitatively demonstrate the impressive results achieved by our proposed method on constructed challenging MPHPE benchmarks. 
\end{itemize}


\section{Related Work}

Head pose estimation is a popular and widely researched field \cite{murphy2008head, khan2021head, abate2022head}. Regardless of its other branches, such as methods based on multi-view or RGB-D inputs, this paper focuses on the HPE task using static monocular RGB images.

\subsection{Head Pose Estimation}

\subsubsection{Traditional Methods}
They include template matching, cascaded detectors, linear regressions and deformable models. Template matching finds the nearest matches by comparing inputs with labeled pose template set \cite{ng2002composite}. Cascaded detectors comprise distinct head detectors attuned to specific poses \cite{murphy2008head}. Linear regressions learn to map high-dimensional feature vectors onto the joint space of head pose angles \cite{drouard2017robust}. These methods are not robust to complex scenes due to the limitations of {\it handcrafted} rules. Deformable models determine the head pose of input images by matching detected facial features including keypoints \cite{kazemi2014one} or landmarks \cite{zhu2016face,bulat2017far} with a static or deformable template (e.g., Active Shape Model \cite{cootes2004statistical}). Their challenge lies in extracting precise facial feature, which is itself a well-studied problem. We recommend readers to refer the survey \cite{murphy2008head, abate2022head} for more details.

\subsubsection{Deep Supervised Learning}\label{HPE_DSL}
Benefiting from advanced deep neural networks, data-driven supervised learning approaches tend to dominate the HPE field in recent years. Basically, we can divide them into three categories: landmark-based, landmark-free and 3DMM-based.

{\bf Landmark-based:} These methods first detect the sparse or dense face landmarks, and then estimate the head pose according to these keypoints, e.g., applying solvePnP \cite{bradski2000opencv}. Although CNN-based face alignment methods \cite{wu2019facial} are accurate enough, such as Dlib \cite{kazemi2014one}, FAN \cite{bulat2017far}, 3DDFA \cite{zhu2016face} and 3DDFA\_v2 \cite{guo2020towards}, they may suffer from high computational costs and inevitable difficulties involving large-angle, partial occlusion or low resolution. Once the detected landmarks are in a form of chaos, the accuracy of head pose will be seriously impaired. Therefore, researches often avoid utilizing the generated landmarks directly. For example, KEPLER \cite{kumar2017kepler} presents an iterative method for joint keypoint estimation and pose prediction of unconstrained faces by its proposed heatmap-CNN architecture. Barra \etal \cite{barra2020web} firstly predicts the positions of 68 well-known face landmarks, and then selects the best exemplar by applying a web-shaped model. EVA-GCN \cite{xin2021eva} constructs a landmark-connection graph, and proposes to leverage the Graph Convolutional Networks (GCN) to model the complex nonlinear mappings between the graph typologies and poses. Similarly, OsGG-Net \cite{mo2021osgg} proposes a one-step graph generation network for estimating head poses by modelling Euler angles associated with the landmark distribution. HHP-Net \cite{cantarini2022hhp} estimates the head pose by a small set of computed head keypoints, and provides a measure of the heteroscedastic uncertainties associated with Euler angles. Nonetheless, landmark-based methods are always inferior to their landmark-free or 3DMM-based counterparts due to its inherent defects.

{\bf Landmark-free:} Instead of relying on facial landmarks, these methods take well-cropped faces as inputs, and directly regress head pose by training on labeled data \cite{ruiz2018fine, hsu2018quatnet, yang2019fsa, zhou2020whenet, zhang2020fdn, valle2020multi, geng2020head, bisogni2021fashe, cao2021vector, albiero2021img2pose}. The mainstream pose representation of single HPE is Euler angles. HopeNet \cite{ruiz2018fine} firstly proposes a multi-loss framework to predict Euler angles from image intensities through joint binned pose classification and regression. Inspired by age estimation method SSR-Net-MD \cite{yang2018ssr}, FSA-Net \cite{yang2019fsa} presents a compact model combining the soft stagewise regression scheme, fine-grained structure features aggregation, and multiple spatial attention mechanism. WHENet \cite{zhou2020whenet} extends the narrow-range HPE method HopeNet \cite{ruiz2018fine} to the full-range of yaws. RAFA-Net \cite{behera2020rotation} automatically captures face details by introducing the self-attention mechanism into the network structure. FDN \cite{zhang2020fdn} claims that the three interrelated Euler angles are mutually exclusive. A feature decoupling network with triple branches is proposed to implicitly constrain the independence of three angles. Focusing on the uncertainty and ambiguity of pose labels, MLD \cite{geng2020head} uses soft labels rather than explicit hard labels to indicate the head pose through its proposed multivariate label distribution. Kuhnke \etal \cite{kuhnke2021relative} proposes a semi-supervised learning strategy for HPE based on relative pose consistency regularization. Recently, LwPosr \cite{dhingra2022lwposr} introduces the depth separable convolution and transformer encoder layers, and maintains relatively high HPE accuracy while reducing network parameters. Besides, some researches exploit the synergy among the HPE-contained multi-task learning which boosts up each individual. Such as HyperFace \cite{ranjan2017hyperface}, MNN \cite{valle2020multi}, MOS \cite{liu2021mos} and SynergyNet \cite{wu2021synergy}.

On the other hand, to avoid the ambiguity problems (e.g., gimbal lock and nonstationary property) of the widely used Euler angles, some alternative representations are adopted. For example, QuatNet \cite{hsu2018quatnet} predicts the quaternion $(x,y,z,w)$ of head pose by designing a new ordinal regression loss based on the characteristics of quaternion $(x^2\!+\!y^2\!+\!z^2\!+\!w^2\!=\!1)$. TriNet \cite{cao2021vector} uses three mutually perpendicular direction vectors in a $\mathbb{R}_{3\times3}$ orthogonal rotation matrix to represent the head pose, and proposes the corresponding orthogonal loss to reduce regression errors. MFDNet \cite{liu2021mfdnet} also applies the rotation matrix representation. It focuses on the ambiguity and uncertainty of the rotation label, and proposes to fit the matrix Fisher distribution of the rotation matrix by an exponential probability density model. Zhou \etal \cite{zhou2022intuitive} explores to adopt the 3D object detection to predict a $\mathbb{R}_{2\times8}$ orthogonally projected head cube, and then converts it into Euler angles. Considering that a rotation matrix needs 9-DoF which is redundant, some methods \cite{dai2020rankpose, albiero2021img2pose, hempel20226d} exploit to compress it by designing simplified vectors. RankPose \cite{dai2020rankpose} reformulates Euler angles by constraining them to a bounded $\mathbb{R}_{1\times3}$ vector space, and represents the head pose as vector projection or vector angles. Img2Pose \cite{albiero2021img2pose} adopts a 6-DoF vector to simultaneously indicate the position and pose information of the face. 6DRepNet \cite{hempel20226d} utilizes two mutually perpendicular unit vectors for achieving easier optimization. The third vector is recovered by applying the Gram-Schmidt orthogonalization.

Generally, these new representations have helped to improve the HPE accuracy to some extent. However, their ground-truths are converted from Euler angles, which are also used for the final error evaluation. Without loss of generality, we follow the frequent used representation Euler angles, and carry forward the full-range angle predictions.  

{\bf 3DMM-based:} In addition to being an independent vision task, HPE can also be attributed as a subtask and included in the regression of 3D Morphable Model (3DMM) parameters, which is originally proposed for the 3D face reconstruction. 3DDFA \cite{zhu2016face} proposes a 3D dense face alignment framework by first introducing the 3DMM fitting of BFM \cite{paysan20093d} integrated into a cascaded-CNN. It conveniently develops a face profiling method to synthesize abundant training samples to solve the rareness of head pose data. 3DDFA\_v2 \cite{guo2020towards} extends it into a more accurate and stable one by proposing a meta-joint optimization strategy. HeadFusion \cite{yu2018headfusion} presents a framework that combines the strengths of a 3DMM model fitted online with a prior-free reconstruction of a 3D full head model which provides support for pose estimation from any viewpoint. SynergyNet \cite{wu2021synergy} combines 3DMM fitting and 3D face landmarks detection tasks together by using a bi-directional optimization process to exploit their synergy. Recently, DAD-3DHeads \cite{martyniuk2022dad} releases a large-scale dense, accurate, and diverse 3D head dataset, labeled with accurate landmarks based on FLAME \cite{li2017learning} model. Then, it provides a baseline DAD-3DNet for the 3D head model parameters regression.

Although 3DMM-based methods have obtained state-of-the-art results on the HPE branch, most of them are limited to a single head with visible face. And the cost of labeling large-scale 3D head ground-truths of arbitrary 2D head images is considerable, which hinders its application in-the-wild.

\subsection{Head Pose Datasets}

\subsubsection{Single Person HPE}
Currently, there are two primary benchmarks: 300W-LP \& AFLW2000 \cite{zhu2017face} and BIWI \cite{fanelli2013random}. 300W-LP and AFLW2000  are created by 3DDFA \cite{zhu2016face} using the 3DMM BFM \cite{paysan20093d} to fit to faces under large pose variation and report Euler angles. BIWI contains 24 videos recorded people sitting in front of the camera and turning their heads. Head poses are estimated by in-depth information from Kinect. A major drawback of these datasets is that the Euler angles are {\it narrowed} in the range of $(-99^{\circ}, 99^{\circ})$. From that disadvantage, for training and validation in full-range HPE, WHENet \cite{zhou2020whenet} designs an auto-labeling process using the CMU Panoptic Dataset \cite{joo2015panoptic} which provides comprehensive yaw angle from $(-180^{\circ}, 180^{\circ})$. Besides, SynHead \cite{gu2017dynamic} constructs a synthetic head dataset for video-based HPE to cope with the need for large training data with accurate annotations. PADACO \cite{kuhnke2019deep} combines BIWI and SynHead to get SynHead++, and presents domain adaptation for HPE with a focus on continuous pose label spaces. 2DHeadPose \cite{wang20232dheadpose} is a new in-the-wild dataset. It is constructed by the proposed annotation method for the head pose in RGB images by using a 3D virtual human head to manually simulate the head pose.

\subsubsection{Multi-Person HPE}
So far, there is no dedicated MPHPE dataset that provides precise head pose labels. In MOS \cite{liu2021mos} and Img2Pose \cite{albiero2021img2pose}, they annotate the dataset WIDER FACE \cite{yang2016wider} with head pose labels in a semi-supervised way, which means skipping tiny faces and auto-labeling by applying existing HPE methods (e.g., FSA-Net \cite{yang2019fsa} or RetinaFace\cite{deng2020retinaface}+PnP). This weakly supervised learning way has much space for improvement, and the labeled WIDER FACE does not contain head samples with invisible faces. Therefore, we put forward new full-range MPHPE benchmarks that support head detection and pose estimation simultaneously from all viewpoints. Following WHENet, yet without cropping out the person head in each image, we automatically extract ground-truth bounding boxes and Euler angles of all heads in images from both datasets: AGORA \cite{patel2021agora} and CMU Panoptic \cite{joo2015panoptic}. More details about processing are in Section \ref{benchmark}.

\subsection{Multi-Person Head Pose Estimation}
In current HPE applications, all faces or heads in images need to be localized formerly by applying off-the-shelf detectors \cite{zhang2016joint, deng2020retinaface, jocher2020yolov5}, which is inherently defective as described in Section \ref{intro}. Although some methods \cite{zhu2012face, chen2014joint, zhang2016joint, ranjan2017hyperface, ranjan2017all, valle2020multi, liu2021mos} combine the single HPE with other facial analysis problems (e.g., face detection and landmark localization) in a multi-task framework, they still take single head image as input. The most relevant work to MPHPE is Img2Pose \cite{albiero2021img2pose}, which achieves joint face detection and pose estimation. However, due to its visible face based optimization design and absence of dedicated full-view training data, it cannot simply migrate to the full-range MPHPE task.

In this paper, inspired by bottom-up multi-person pose estimation OpenPose \cite{cao2017realtime} and SPM \cite{nie2019single}, we propose a novel end-to-end one-stage MPHPE network which supports head detection and pose estimation simultaneously. In pursuit of a balance between accuracy and speed, we adopt the cost-effective one-stage YOLOv5 \cite{jocher2020yolov5} as the detection backbone. YOLO series \cite{redmon2016you, redmon2017yolo9000, bochkovskiy2020yolov4, redmon2018yolov3} have been proven to exceed the justifiable trade-off of their two-stage counterparts \cite{ren2015faster, he2017mask}. By integrating Cross-Stage-Partial (CSP) bottlenecks \cite{wang2020cspnet}, Feature Pyramid Network (FPN) \cite{lin2017feature} and Focal Loss \cite{lin2017focal}, YOLOv5 has surpassed most two-stage and one-stage (e.g., SSD \cite{liu2016ssd} and FCOS \cite{tian2019fcos}) detectors in accuracy and speed.


\section{MPHPE Benchmarks}\label{benchmark}

We construct two MPHPE benchmarks by building human head detection ({\it 2D bounding box}) datasets, and extracting the corresponding head orientation ({\it three Euler angles}) as an attribute label. Some examples are in Fig.~\ref{datasetexamples}.

\subsection{Benchmarks Construction}

\subsubsection{AGORA-HPE}
AGORA is a synthetic dataset based on realistic 3D  environments and SMPL-X \cite{pavlakos2019expressive} body models, originally for the 3D Human Pose and Shape (3DHPS) estimation task. Its released data provides rich labels on images, including 2D person masks, SMPL-X ground-truth fittings and camera information\footnote{https://agora.is.tue.mpg.de/}. To generate MPHPE labels for each head, we further extract camera parameters $\mathbf{C}_{real}$ and 3D face landmarks as heads $\mathcal{H}_{real}$.

Specifically, for each given head $\mathcal{H}_{real}\in\mathbb{R}_{3\times N}$ with $N$ 3D face landmarks, we firstly compute the similarity transformation matrix $\mathbf{M}_c$ from a generic head model $\mathcal{H}_{ref}$ with an exactly frontal view and pre-defined camera parameters $\mathbf{C}_{ref}$ by the closed-form solution \cite{horn1987closed}. For fast and robust computation, we elaborately select $N'$ pairs of corner landmarks in $\mathcal{H}_{real}$ and $\mathcal{H}_{ref}$ to align. Then, we define a transformed 3D hemisphere loosely around each head using $\mathbf{M}_c$, and generate its bounding box by the 2D projection with real camera parameters $\mathbf{C}_{real}$. Instead of projecting face landmarks, this way can obtain {\it an area containing the background and whole head}. A loose head bounding box is proved to be more beneficial than a tight one in \cite{shao2019improving}. Finally, to extract the head orientation, we estimate the transformation matrix $\mathbf{M}_r$ from the camera world $\mathbf{C}_{cam}$ to the real world $\mathbf{C}_{real}$. The $\mathbf{C}_{cam}$ is obtained by $\mathbf{C}_{ref}{\mathbf{M}_c^{-1}}$. We calculate $\mathbf{M}_r$ as below:
\begin{equation}
	\mathbf{M}_r = \mathbf{C}_{real}{\mathbf{C}_{cam}^{-1}} = \mathbf{C}_{real}{\mathbf{M}_c}{\mathbf{C}_{ref}^{-1}} ~
\end{equation}
with the recovered $\mathbf{M}_r$, we then split out three Euler angles in the {\it pitch-yaw-roll} order following datasets 300W-LP \cite{zhu2017face} and BIWI \cite{fanelli2013random}.

Although AGORA is designed to place 5$\sim$15 human bodies in each image, due to the occlusion and truncation of heads, some images may not have at least one valid head pose label and are therefore discarded. The final generated AGORA-HPE benchmark contains {\bf 1,070} and {\bf 14,408} images for validation and train sets, respectively.

\subsubsection{CMU-HPE}
\begin{figure}[!t]
\centering
\includegraphics[width=\columnwidth]{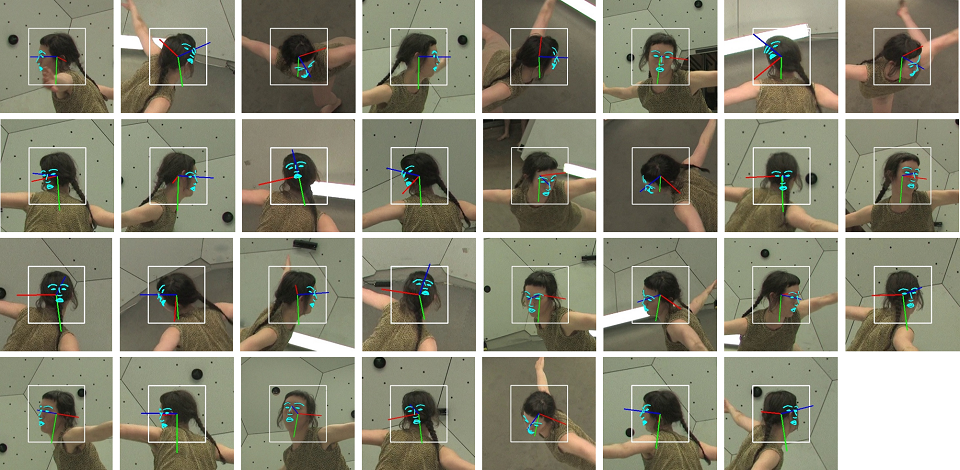}
\caption{Snapshot of 31 views from the sequence $\mathsf{170307\_dance5}$ at sampling moment $T\!=\!24$ seconds. Frames are cropped for ease of presentation.}
\label{snapshot}
\end{figure}

The CMU Panoptic Dataset is collected by a massive multi-view system. Its scenes focus primarily on a single person or interacted people in a hemispherical device. And its labels include 3D body poses, hand keypoints and facial landmarks of multiple people in 31 synchronized HD video streams. It also provides calibrated camera parameters $\mathbf{C}_{real}$ from 31 views. From 84 different topics published on its official website\footnote{http://domedb.perception.cs.cmu.edu/}, we select 17 topics and uniformly sample frames every 2 or 6 seconds from 31 HD videos under each topic. A typical sampling moment snapshot is shown in Fig.~\ref{snapshot}. Then, similar to the process of building AGORA-HPE, we construct the CMU-HPE which has {\bf 16,216} and {\bf 15,718} images for validation and train sets, respectively. 

\subsection{Data Distribution}

We compare our two constructed benchmarks AGORA-HPE and CMU-HPE with existing commonly used single HPE datasets: 300W-LP\&AFLW2000 and BIWI. The 300W-LP includes 122,450 training images, and AFLW2000 includes 2,000 testing images. In two datasets, yaw angle is in the range of $(-99^{\circ}, 99^{\circ})$, pitch and roll angles are in the range $(-90^{\circ}, 90^{\circ})$. The BIWI has over 15,000 images. The head pose covers $(-75^{\circ}, 75^{\circ})$ of yaw angle and $(-60^{\circ}, 60^{\circ})$ of pitch and roll angles.

\subsubsection{Angles Range}
\begin{figure}[!t]
\centering
\includegraphics[width=\columnwidth]{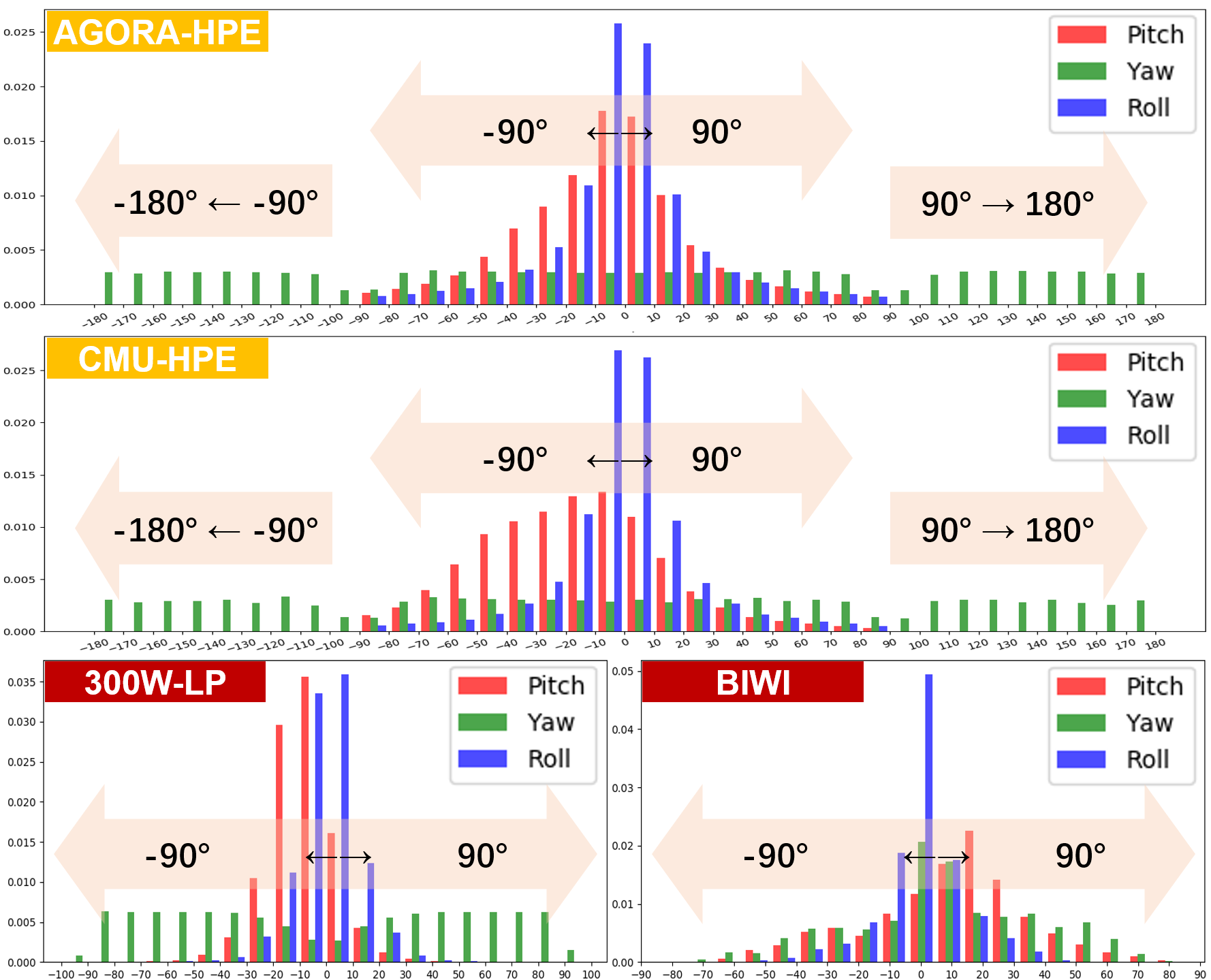}
\caption{Pose label distribution of three Euler angles in datasets AGORA-HPE, CMU-HPE, 300W-LP\&AFLW2000 and BIWI.}
\label{statistic}
\end{figure}

The distribution of Euler angles in all datasets is shown in Fig.~\ref{statistic}. Basically, the pitch and roll angles follow a normal distribution. However, our two datasets visually have higher variance, which indicates their rich pose diversity. The yaw angles are roughly uniformly distributed except for the simplest BIWI dataset. In addition to having full-range yaw angles, AGORA-HPE and CMU-HPE are datasets for the more challenging MPHPE task, with about {\bf 7.27} and {\bf 2.14} heads per image, respectively.

\subsubsection{Pose Label Variance}
To evaluate the pose label quality of two newly built datasets, we chose to measure the standard deviation of all head pose obtained using different number of $N$ referred 3D face landmarks. By selecting $N$ from the set $\{9, 11, 13, 15, 17\}$, we then computed and obtained the standard deviation of $(pitch, yaw, roll, avg)$  as ${\bf (2.29, 0.57, 1.73, 1.91)}$ and ${\bf (2.13, 0.79, 1.65, 1.81)}$ on datasets AGORA-HPE and CMU-HPE, respectively. These low variance values demonstrate relative reliability and insignificant divergence of the generated pose labels between different head aligning ways. For another reference, the errors reported by the SOTA method 6DRepNet \cite{hempel20226d} on the easiest single HPE dataset BIWI are ${\bf (2.92, 2.69, 2.36, 2.66)}$, which is still far from our label variance. We finally set $N=13$ for pose label generation of both datasets.

\subsubsection{Challenging Samples}
\begin{figure}[!t]
\centering
\includegraphics[width=\columnwidth]{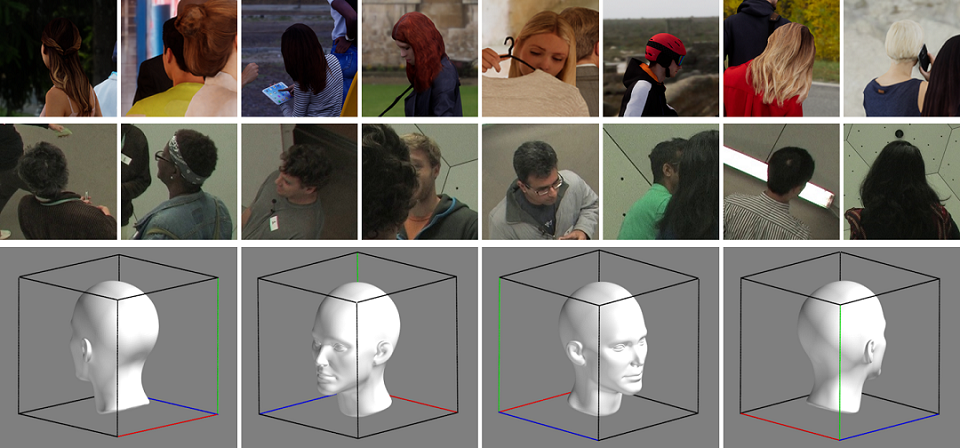}
\caption{Example of some challenging head samples from AGORA-HPE ({\it first line}) and CMU-HPE ({\it second line}). The {\it third line} is 3D head model indicator.}
\label{hardsamples}
\end{figure}

Generally, our proposed two full-range datasets naturally contain more face invisible heads than 300W-LP\&AFLW2000 and BIWI. As shown in Fig.~\ref{hardsamples}, except for those peculiar backward heads, ordinary frontal faces are often co-existed with complex conditions such as self-occlusion, emerged-occlusion or abnormal angles. Accurate and robust pose estimation of these challenging full-range heads needs to rely on more background areas or the context of related upper human bodies, which cannot be well-addressed by current single HPE researches. 

\section{Proposed Direct MPHPE}\label{approach}

In this section, we describe our method DirectMHP towards joint head detection and pose estimation of multiple persons.

\subsection{Merging: Joint Detection and Estimation}  

In our proposed approach to MPHPE, we train a dense detection network to directly predict a set of head objects $\{\widehat{\mathcal{O}}\in\widehat{\mathbf{O}} \parallel \widehat{\mathcal{O}}=cat(\widehat{\mathcal{O}}^b,\widehat{\mathcal{O}}^p), \widehat{\mathbf{O}}=\widehat{\mathbf{O}}^b \cup \widehat{\mathbf{O}}^p\}$, which contains head bounding boxes set $\widehat{\mathbf{O}}^b$ and corresponding head pose set $\widehat{\mathbf{O}}^p$ concurrently. We briefly present the meaning of each notation, and exploit the inherent correlation between joint head detection and pose estimation.

Actually, one head object $\mathcal{O}$ is essentially an extension of the conventional offset-based object representation $\mathcal{O}^b$ which locates an object with a tight bounding box $\mathbf{b}=(b_x, b_y, b_w, b_h)$ around it. The coordinates $(b_x, b_y)$ are the center position of $\mathbf{b}$. The $b_w$ and $b_h$ are the width and height of $\mathbf{b}$, respectively. For the head pose representation $\mathcal{O}^p$, we adopt the commonly used three Euler angles $\mathbf{p}=(p_{pitch}, p_{yaw}, p_{roll})$. Then, instead of following the incompact multi-task cascaded CNN framework (e.g., MTCNN \cite{zhang2016joint}), we treat head pose $\mathcal{O}^p$ as an additional head attribute, and concatenate it with its head position $\mathcal{O}^b$ to build a joint representation of the head object $\mathcal{O}$. We then integrate these two tasks together by using the joint prediction in an unified framework.

Intuitively, we can benefit a lot from joint representation and prediction. On one hand, an appropriate head bounding box $\mathcal{O}^b$ possesses both strong local characteristics (e.g., eyes, ears and jaws) and weak global features (such as surrounding background and anatomical position) for its head orientation $\mathcal{O}^p$ estimation. Thus, we bind the $\mathcal{O}^b$ and $\mathcal{O}^p$ into one embedding to enable the network to learn their intrinsic relationships. One the other hand, compared to methods using multiple tasks or stages, mixing $\mathcal{O}^b$ and $\mathcal{O}^p$ up can be leveraged directly in MPHPE without the need of hand-crafted post-processing or parsing. By designing a one-stage network DirectMHP that uses shared heads to jointly predict $\mathcal{O}^b$ and $\mathcal{O}^p$, our approach can achieve high accuracy with minimal computational burden during training and inference.

\subsection{Network Architecture Design}\label{netarch}

\begin{figure*}[t]
	\centering
	\subfloat[Overall network structure]{\includegraphics[width=0.416\textwidth]{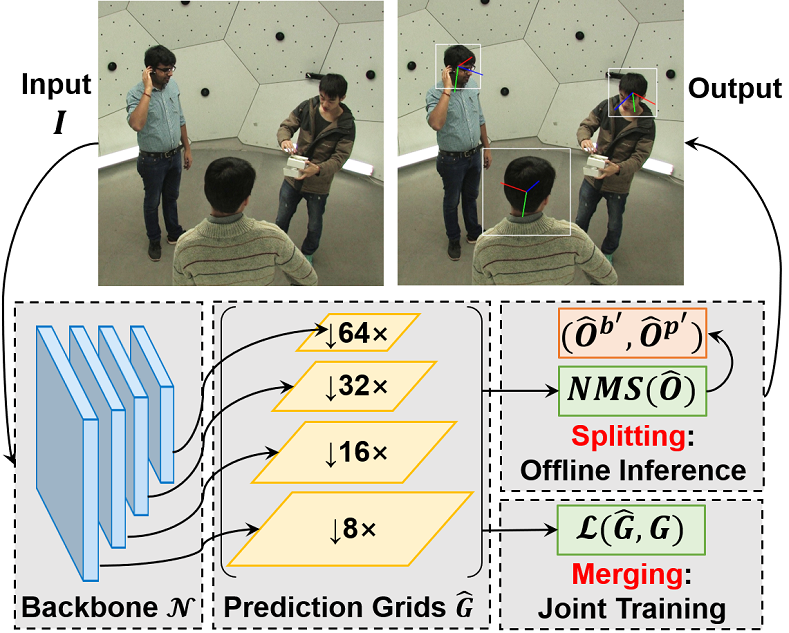}
	\label{networka}}
	\hfill
	\subfloat[Illustration of grid cell predictions]{\includegraphics[width=0.554\textwidth]{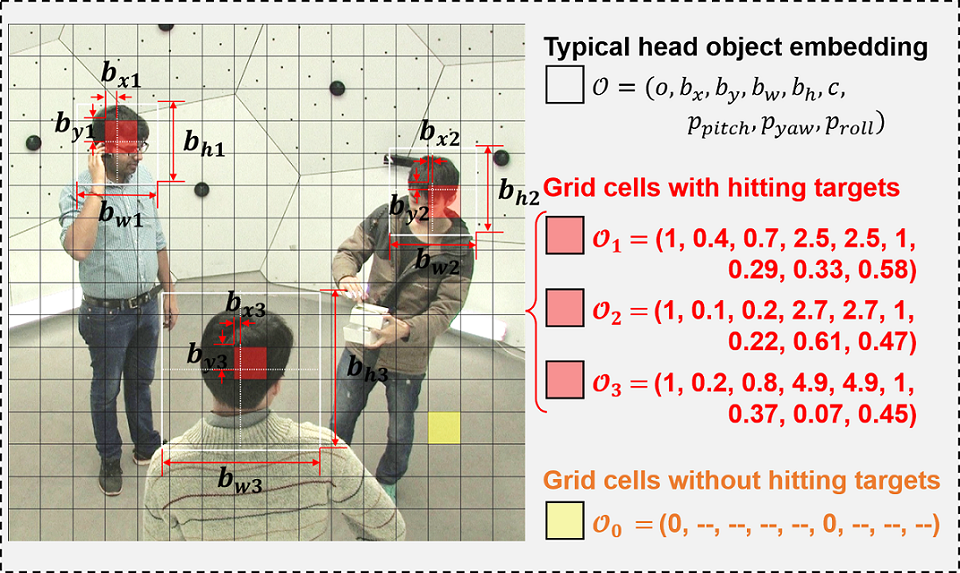}%
	\label{networkb}}
	\caption{{\bf (a)} Our DirectMHP adopts an object detection network $\mathcal{N}$ (e.g., YOLOv5) as the backbone to extract features and generate prediction grids $\mathit{\widehat{G}}$ from one input image $\mathbf{I}$. During training, we use target grids $\mathit{G}$ to supervise the loss function $\mathcal{L}$. In inference, we apply the Non-Maximum Suppression (NMS) on predicted head objects $\widehat{\mathbf{O}}$ to obtain final bounding boxes set $\widehat{\mathbf{O}}^{b'}$ and related head pose set $\widehat{\mathbf{O}}^{p'}$. {\bf (b)} In the illustration example for grid cells, these with hitting head objects have red color and without hitting targets have yellow color. The “--” is not used when calculating loss.}
	\label{network}
\end{figure*}

Our network structure is shown in Fig.~\ref{networka}. We adopt an object detection architecture $\mathcal{N}$ as the basic backbone. Specifically, we choose the recently most cost-effective one-stage YOLOv5 \cite{jocher2020yolov5} as the feature extractor. We do not illustrate all details of $\mathcal{N}$ for simplicity. Following YOLOv5, we feed $\mathcal{N}$ one RGB image $\mathbf{I}\in\mathbb{R}^{h\times w\times 3}$ as the input, and output four multi-scale grids $\widehat{\mathit{G}}=\{\widehat{\mathcal{G}}^s \parallel s\in\{8,16,32,64\}\}$ from four prediction heads. Each grid $\widehat{\mathcal{G}}\in\mathbb{R}^{C_a\times C_o\times \frac{h}{s}\times \frac{w}{s}}$ contains dense head object outputs $\widehat{\mathbf{O}}$ produced from $C_a$ anchor channels and $C_o$ output channels.

For the anchor-based method YOLOv5, we generally predict grids $\widehat{\mathit{G}}$ in different levels decided by the sliding of convolution kernels, the scale of feature maps and the definition of anchors. Supposing that one target object $\mathcal{O}$ is centered at $(b_x, b_y)$ in the feature map $\mathbf{F}^s$, the corresponding grid $\widehat{\mathcal{G}}^s$ at cell $(\frac{b_x}{s}, \frac{b_y}{s})$ should be highly confident. When having defined $C_a$ anchor boxes $\mathcal{B}^s=\{(B^w_i,  B^h_i)\vert^{C_a}_i\}$ for the grid $\widehat{\mathcal{G}}^s$, we will thus generate $C_a$ anchor channels at each cell $(\frac{b_x}{s}, \frac{b_y}{s})$. Furthermore, to obtain robust capability, we allow detection redundancy of multiple objects and four surrounding cells matching for each cell. However, this redundancy makes sense to the detection of head position $\mathcal{O}^b$, it is not facilitative to the estimation of head pose $\mathcal{O}^p$. We interpret how to suppress this problem in Section \ref{loss}.

Then, we explain the arrangement of one output object $\widehat{\mathcal{O}}$ from $C_o$ output channels of $\widehat{\mathcal{G}}^s_{i,x,y}$ which is related to $i$-th anchor box at grid cell $(x,y)$. As shown in the example of grid cells in Fig.~\ref{networkb}, one typical predicted $\widehat{\mathcal{O}}$ embedding consists of four parts: the objectness or probability $\hat{o}$ that a head object exists, the candidate bounding box $\mathbf{\hat{b}}'=({\hat{b}_x}', {\hat{b}_y}', {\hat{b}_w}', {\hat{b}_h}')$, the head object score $\hat{c}$, and the candidate Euler angles $\mathbf{\hat{p}}'=({\hat{p}_{pitch}}', {\hat{p}_{yaw}}', {\hat{p}_{roll}}')$. Thus, $C_o=9$. Moreover, to transform the candidate $\mathbf{\hat{b}}'$ into coordinates $\mathbf{\hat{b}}$ relative to the grid cell $\widehat{\mathcal{G}}^s_{i,x,y}$, we use strategies as below:
\begin{align}
	&\enspace{\hat{b}_x}=2\phi({\hat{b}_x}')-0.5, \qquad {\hat{b}_y}=2\phi({\hat{b}_y}')-0.5 ~\\
	&{\hat{b}_w}=\frac{B^w_i}{s}[2\phi({\hat{b}_w}')]^2, \qquad {\hat{b}_h}=\frac{B^h_i}{s}[2\phi({\hat{b}_h}')]^2 ~
\end{align}
where $\phi$ is the sigmoid function that limits model predictions in the range $(0,1)$. Similarly, we obtain $\mathbf{\hat{p}}$ by applying $\phi(\cdot)$ on candidate angle $\mathbf{\hat{p}}'$ and rescaling them to real degrees:
\begin{align}
	&{\hat{p}_{yaw}}=[\phi({\hat{p}_{yaw}}')-0.5] \times 360^{\circ}  \\
	&{\hat{p}_{pitch}}=[\phi({\hat{p}_{pitch}}')-0.5] \times 180^{\circ} \\
	&{\hat{p}_{roll}}=[\phi({\hat{p}_{roll}}')-0.5] \times 180^{\circ} ~
\end{align}
where $\hat{p}_{yaw}$ has full-range angles $(-180^{\circ}, 180^{\circ})$. The $\hat{p}_{pitch}$ and $\hat{p}_{roll}$ are all in range $(-90^{\circ}, 90^{\circ})$. To learn $\mathbf{\hat{p}}$ better during training, we will keep using $\mathbf{\hat{p}'}$ with values in the range $(0,1)$ for its loss calculation.

\subsection{Multi-Loss Optimization}\label{loss}

For a set of predicted grids $\mathit{\widehat{G}}$, we firstly build the corresponding target grids set $\mathit{G}$ following formats introduced in Section \ref{netarch}. Then, we calculate the total training loss $\mathcal{L}=(\mathit{\widehat{G}},\mathit{G})$ as follows:
\begin{equation}
	\mathcal{L}=N_{bs}(\alpha\mathcal{L}_{box} + \beta\mathcal{L}_{obj} + \gamma\mathcal{L}_{mse}) ~
	\label{losstotal}
\end{equation}
where $N_{bs}$ is the training batch size. The $\alpha$, $\beta$ and $\gamma$ are weights of losses $\mathcal{L}_{box}$, $\mathcal{L}_{obj}$ and $\mathcal{L}_{mse}$, respectively. Because only having a sinlge head object score $\hat{c}$, we do not need the classification loss $\mathcal{L}_{cls}$ as in YOLOv5. The $\hat{c}$ will be used to compute the object confidence. We compute three loss components as follows:
\begin{align}
	&\mathcal{L}_{box}=\sum\nolimits_s\frac{1}{\|\mathcal{G}^s\|}\sum\nolimits^{\|\mathcal{G}^s\|}_{i=1}[1-\mathsf{CIoU}(\mathbf{\hat{b}_i}, \mathbf{b_i})] \\~
	&\mathcal{L}_{obj}=\sum\nolimits_s\frac{w_s}{\|\mathcal{G}^s\|}\sum\nolimits^{\|\mathcal{G}^s\|}_{i=1}\mathsf{BCE}(\hat{o}, o\cdot\mathsf{CIoU}(\mathbf{\hat{b}_i}, \mathbf{b_i})) \\~
	&\mathcal{L}_{mse}=\sum\nolimits_s\frac{1}{\|\mathcal{G}^s\|}\sum\nolimits^{\|\mathcal{G}^s\|}_{i=1}\varphi(\hat{o}>\tau)\| \mathbf{\hat{p}'_i}-\mathbf{p'_i} \|_2 ~
	\label{lossmse}
\end{align}

For the bounding box regression loss $\mathcal{L}_{box}$, we adopt the complete intersection over union (CIoU) \cite{zheng2020distance} across four grids $\mathcal{G}^s$. $\mathsf{BCE}$ in the objectness loss $\mathcal{L}_{obj}$ is the binary cross-entropy. The multiplier $o$ in it is used for penalizing candidates without hitting target grid cells ($o=0$), and encouraging candidates around target anchor ground-truths ($o=1$). The $w_s$ is a balance weight for different grid level. Finally, we utilize the mean squared error (MSE) to measure original outputs $\mathbf{\hat{p}'}$ and normalized targets $\mathbf{p'}$ in loss $\mathcal{L}_{mse}$. Before that, we apply a tolerance threshold $\tau$ on objectness $\hat{o}$ to filter out those false-positive redundant predictions from $\widehat{\mathcal{O}}$. These lower-scoring proposed regions may have less or no human head information. They are intuitively harmful to $\mathcal{L}_{mse}$.

Besides, WHENet \cite{zhou2020whenet} has pointed out the over-penalizing ($>180^{\circ}$) problem for the full-range yaw angles predictions. And it defines a new wrapped MSE loss by choosing the minimal rotation angle to align each yaw prediction:
\begin{equation}
    \mathcal{L}_{wmse}=\frac{1}{N}\sum^N_{i=1}\mathsf{min}(\| \mathbf{\hat{p}'_i}-\mathbf{p'_i} \|_2, \|1-|\mathbf{\hat{p}'_i}-\mathbf{p'_i}|\|_2) ~
\end{equation}
where $|\mathbf{\hat{p}'_i}-\mathbf{p'_i}|$ is the absolute error of angles. For a pair of normalized yaw angles $(\hat{p}'_{yaw}, {p}'_{yaw})$, the absolute error should not be greater than 0.5 ($>180^{\circ}$) due to the closed rotation of heads. However, we declare that it is not necessarily reasonable to replace $\mathcal{L}_{mse}$ with $\mathcal{L}_{wmse}$ in training (see Section \ref{sectionAS}). We attribute this to the inherent ability of deep models to learn to approximate the true yaw angle from a single direction. Nonetheless, we should still follow this truth when evaluating yaw predictions in Section \ref{evaluation}.

\subsection{Splitting: Offline Inference}

After training, we need to process the predicted objects set $\widehat{\mathbf{O}}$ to get final results. First of all, we apply the conventional Non-Maximum Suppression (NMS) to filter out false positives and redundant bounding boxes:
\begin{equation}
	(\widehat{\mathbf{O}}^{b'}, \widehat{\mathbf{O}}^{p'})=\mathsf{NMS}(\widehat{\mathbf{O}}, \tau_{conf}, \tau_{iou}) ~
\end{equation}
where $\tau_{conf}$ and $\tau_{iou}$ are thresholds for object confidence and IoU overlap, respectively. We calculate confidence of each predicted object $\widehat{\mathcal{O}}$ by $\hat{o}\cdot\hat{c}$. We do not need to modify the common $\mathsf{NMS}$ steps of getting positive head bounding boxes $\widehat{\mathbf{O}}^{b'}$ which are accompanied by head orientations $\widehat{\mathbf{O}}^{p'}$. Then, we rescale $\widehat{\mathbf{O}}^{b'}$ and $\widehat{\mathbf{O}}^{p'}$ to obtain real $\widehat{\mathbf{O}}^{b}$ and $\widehat{\mathbf{O}}^{p}$ by following transformations:
\begin{align}
	&\widehat{\mathcal{O}}^b=s\cdot[\widehat{\mathcal{O}}^{b'}+(x_o, y_o, 0, 0)] \\~
	&\widehat{\mathcal{O}}^p=(\widehat{\mathcal{O}}^{p'}-0.5)\times L ~
	\label{poseback}
\end{align}
where $x_o$ and $y_o$ are offsets from grid cell centers. The $s$ maps the box size back to the original image shape. The $L$ means the range of three Euler angles. We finally report all evaluation results on $\widehat{\mathbf{O}}^{b}$ and $\widehat{\mathbf{O}}^{p}$.


\section{Experiments}

\subsection{Training Details}
We adopt the PyTorch 1.10 and 4 RTX-3090 GPUs for implementation. Following the original YOLOv5 \cite{jocher2020yolov5}, we train two kinds of models including DirectMHP-S and DirectMHP-M by controlling the depth and width of CSP bottlenecks \cite{wang2020cspnet} in $\mathcal{N}$. Firstly, to compare with the dominant single HPE methods, following the {\bf protocol 1} in FSA-Net \cite{yang2019fsa}, we train DirectMHP-M 300 epochs on 300W-LP with input shape $512\times512\times3$, and test on val-sets AFLW2000 and BIWI. For the MPHPE task, depending on the dataset complexity, we train on AGORA-HPE and CMU-HPE for 300 and 200 epochs using the SGD optimizer, respectively. The shape of the input image is resized and zero-padded to $1280\times1280\times3$. Besides, for properly comparing with Img2Pose \cite{albiero2021img2pose}, we also pretrain our method in the weakly supervised way on relabeled WIDER FACE dataset by RetinaFace \cite{deng2020retinaface}, and finetune it on the 300W-LP dataset as in \cite{albiero2021img2pose}.

Unlike conventional detection tasks, due to the irreversibility of 2D Euler angles, we cannot randomly augment the dataset through rotation or affine deformation during training and inference in DirectMHP. Thus, we leave out affine-related training data augmentations or test time augmentation (TTA). As for many hyperparameters, we keep most of them unchanged, including adaptive anchors boxes $\mathcal{B}^s$, the grid balance weight $w_s$, and the loss weights $\alpha=0.05$ and $\beta=0.7$. Except for two new hyperparameters $\gamma$ and $\tau$, which are studied in Section \ref{sectionAS}, we set $\gamma=0.1$ and $\tau=0.4$ based on experiments. When inference, we keep using the default thresholds $\tau_{conf}=0.7$ and $\tau_{iou}=0.65$ when applying $\mathsf{NMS}$ on $\widehat{\mathbf{O}}$.

\subsection{Evaluation Metrics}\label{evaluation}
We mainly adopt two kinds of evaluation metrics: the average precision (AP) (e.g., the AP$^{.5:.95}$ defined in the COCO API\footnote{https://cocodataset.org/\#detection-eval}) for head bounding boxes detection, and the mean absolute error (MAE) for three Euler angles estimation. The MAE metric is defined as follows:
\begin{equation}
    MAE=\frac{1}{\hat{n}}\sum^{\hat{n}}_{i=1}\mathsf{min}(| \mathbf{\hat{p}_i}-\mathbf{p_i} |, 360^{\circ}-|\mathbf{\hat{p}_i}-\mathbf{p_i}|) ~
\end{equation}
where $\mathbf{\hat{p}_i}$ and $\mathbf{p_i}$ both contain angles in degrees format. $\hat{n}$ is the number of detected head instances. Considering that not all ground-truths (e.g., totally $n$ heads) will be found in some cases, we only perform IoU matching on each detected head box and compute their MAE. The IoU threshold is set to 0.5. We report $P_M=\hat{n}/n$ to reflect the matching accuracy.

\subsection{Quantitative Comparison}\label{quantitative}
Firstly, we compare our DirectMHP-M directly trained on 300W-LP or pretrained on relabeled WIDER FACE with SOTA HPE methods. Following \cite{ruiz2018fine, yang2019fsa, albiero2021img2pose}, we remove images with yaw angles out of the range ($-99^{\circ}$, $99^{\circ}$) when testing in AFLW2000. Then, on our reconstructed two MPHPE benchmarks, we compare our models quantitatively with four representative landmark-free methods HopeNet \cite{ruiz2018fine}, FSA-Net \cite{yang2019fsa}, Img2Pose \cite{albiero2021img2pose} and 6DRepNet \cite{hempel20226d}. Especially, for FSA-Net\footnote{https://github.com/shamangary/FSA-Net} (Euler angles-based) and 6DRepNet\footnote{https://github.com/thohemp/6DRepNet} (vectors-based), we will retrain them on our benchmarks according to their released code. We also choose four 3DMM-based methods (3DDFA \cite{zhu2016face}, 3DDFA\_v2 \cite{guo2020towards}, SynergyNet \cite{wu2021synergy} and DAD-3DNet \cite{martyniuk2022dad}) which are not convenient to be retrained. For all chosen methods except the retrained full-range 6DRepNet, we compare narrow yaw angles ($|p_{yaw}|<90^{\circ}$) with them. When testing, inputs of all compared methods come from already matched human heads mentioned in Section \ref{evaluation}. We choose head results of DirectMHP-M. Besides, we measure the average inference latency in the same environment to reflect the efficiency of our one-stage approach.


\setlength{\tabcolsep}{5pt}
\begin{table*}[!t]\scriptsize 
	\begin{center}
	\caption{Comparison of our DirectMHP-M with SOTA single HPE methods trained on the 300W-LP dataset. Extra dataset means using additional training data. Red means the best result, and blue means the second-best result.}
	\label{tabSingleHPE}
	\begin{tabular}{l|c|cccc|rrrr|rrrr}
        \Xhline{1.2pt}
        \multirow{2}{*}{\bf Method} & \multirow{2}{*}{\bf Year} & \multirow{2}{*}{\bf Extra dataset?} & \multirow{2}{*}{\makecell{\bf Full-view\\ \bf design?}} & \multirow{2}{*}{\makecell{\bf Multi-\\ \bf tasks?}} & \multirow{2}{*}{\makecell{\bf Head pose\\ \bf representation?}} & \multicolumn{4}{c|}{\bf  \textit{val-set} AFLW2000 \cite{zhu2017face}} & \multicolumn{4}{c}{\bf  \textit{val-set} BIWI \cite{fanelli2013random}} \\
        \cline{7-14}
        {} & {} & {} & {} & {} & {} & {\bf Yaw} & {\bf Pitch} & {\bf Roll} & {\bf MAE} & {\bf Yaw} & {\bf Pitch} & {\bf Roll} & {\bf MAE}   \\
        \Xhline{1.2pt} \rowcolor{gray!15}
        Dlib \cite{kazemi2014one} & 2014 & No & \XSolidBrush & \XSolidBrush & Landmarks (68) & 18.273 & 12.604 & 8.998 & 13.292 & 16.756 & 13.802 & 6.190 & 12.249 \\
        \hline \rowcolor{gray!15}
        FAN \cite{bulat2017far} & 2017 & No & \XSolidBrush  & \XSolidBrush & Landmarks (12) & 6.358 & 12.277 & 8.714 & 9.116 & 8.532 & 7.483 & 7.631 & 7.882 \\
        \hline \rowcolor{gray!15}
        RetinaFace \cite{deng2020retinaface} & 2020 & WIDER FACE & \XSolidBrush & \Checkmark & Landmarks (5) & 5.101 &  9.642 & 3.924 & 6.222 & 4.070 & 6.424 & 2.974 & 4.490  \\
        \hline \rowcolor{gray!15}
        OsGG-Net \cite{mo2021osgg} & 2021 & No & \XSolidBrush  & \XSolidBrush & Landmarks (68) & 3.960 & 5.710 & 3.510 & 4.390 & 3.260 & 4.850 & 3.380 & 3.830 \\
        \hline \rowcolor{gray!40}
        QuatNet \cite{hsu2018quatnet} & 2018 & No & \XSolidBrush & \XSolidBrush & Quaternion & 3.973 & 5.615 & 3.920 & 4.503 & 4.010 & 5.492 & 2.936 & 4.146 \\
        \hline \rowcolor{gray!15}
        HopeNet \cite{ruiz2018fine} & 2018 & No & \XSolidBrush & \XSolidBrush & Euler angles & 6.470 & 6.560 & 5.440 & 6.160  & 4.810 & 6.606 & 3.269 & 4.895 \\
        \hline \rowcolor{gray!15}
        FSA-Net \cite{yang2019fsa} & 2019 & No & \XSolidBrush & \XSolidBrush & Euler angles & 4.501 & 6.078 & 4.644 & 5.074 & 4.560 & 5.210 & 3.070 & 4.280 \\
        \hline \rowcolor{gray!15}
        FDN \cite{zhang2020fdn} & 2020 & No & \XSolidBrush & \XSolidBrush & Euler angles & 3.780 & 5.610 & 3.880 & 4.420 & 4.520 & 4.700 & \color{blue}{\bf 2.560} & 3.930 \\
        \hline \rowcolor{gray!15}
        WHENet \cite{zhou2020whenet} & 2020 & CMU-Panoptic & \Checkmark & \XSolidBrush & Euler angles  & 4.440 & 5.750 & 4.310 & 4.830 & 3.600 & 4.100 & 2.730 & 3.480  \\
        \hline \rowcolor{gray!15}
        MNN \cite{valle2020multi} & 2021 & No & \XSolidBrush & \Checkmark & Euler angles & 3.340 & 4.690 & 3.480 & 3.830 & 3.980 & 4.610 & \color{red}{\bf 2.390} & 3.660 \\
        \hline \rowcolor{gray!15}
        2DHeadPose \cite{wang20232dheadpose} & 2022 & 2DHeadPose & \XSolidBrush & \XSolidBrush & Euler angles  & \color{red}{\bf 2.850} & \color{blue}{\bf 4.470} & \color{blue}{\bf 2.820} & \color{blue}{\bf 3.380} & 3.810 & \color{blue}{\bf 3.780} & 2.730 & \color{red}{\bf 3.440}  \\
	  \hline \rowcolor{gray!40}
        Rankpose \cite{dai2020rankpose} & 2020 & No & \XSolidBrush & \XSolidBrush & Vectors ($1\times3$) & 2.990 & 4.750 & 3.250 & 3.660 & 3.590 & 4.770 & 2.760 & 3.710  \\
        \hline \rowcolor{gray!40}
        TriNet \cite{cao2021vector} & 2021 & No & \Checkmark & \XSolidBrush & Vectors ($3\times3$) & 4.198 & 5.767 & 4.042 & 4.669 & \color{red}{\bf 3.046} & 4.758 & 4.112 & 3.972 \\
        \hline \rowcolor{gray!40}
        MFDNet \cite{liu2021mfdnet} & 2021 & No & \XSolidBrush & \XSolidBrush & Vectors ($3\times3$) & 4.300 & 5.160 & 3.690 & 4.380 & 3.400 & 4.680 & 2.770 & 3.620 \\
        \hline \rowcolor{gray!40}
        Img2Pose \cite{albiero2021img2pose} & 2021 & WIDER FACE & \XSolidBrush & \Checkmark & Vectors ($1\times6$) & 3.426 & 5.034 & 3.278 & 3.913 & 4.567 & \color{red}{\bf 3.546} & 3.244 & 3.786 \\
        \hline \rowcolor{gray!40}
        6DRepNet \cite{hempel20226d} & 2022 & No & \Checkmark & \XSolidBrush & Vectors ($2\times3$) & 3.630 & 4.910 & 3.370 & 3.970 & \color{blue}{\bf 3.240} & 4.480 & 2.680 & \color{blue}{\bf 3.470} \\
        \hline \rowcolor{gray!15}
        3DDFA \cite{zhu2016face} & 2016 & No & \XSolidBrush & \Checkmark & 3DMM & 4.330 & 5.980 & 4.300 & 4.870 & --- & --- & --- & --- \\
        \hline \rowcolor{gray!15}
        3DDFA\_v2 \cite{guo2020towards} & 2020 & No & \XSolidBrush & \Checkmark & 3DMM & 4.060 & 5.260 & 3.480 & 4.270 & --- & --- & --- & --- \\
        \hline \rowcolor{gray!15}
        SynergyNet \cite{wu2021synergy} & 2021 & No & \XSolidBrush & \Checkmark & 3DMM & 3.420 & \color{red}{\bf 4.090} & \color{red}{\bf 2.550} & \color{red}{\bf 3.350} & --- & --- & --- & --- \\ 
        \hline \rowcolor{gray!15}
        DAD-3DNet \cite{martyniuk2022dad} & 2022 & DAD-3DHeads & \Checkmark & \Checkmark & 3DMM & 3.080 & 4.760 & 3.150 & 3.660 & 3.790 & 5.240 & 2.920 & 3.980 \\
        \Xhline{1.2pt}
        DirectMHP-M & 2023 & No & \Checkmark & \Checkmark & Euler angles & 2.988 & 5.351 & 3.767 & 4.035 & 3.566 & 5.475 & 4.022 & 4.355 \\
        \hline
        DirectMHP-M & 2023 & WIDER FACE & \Checkmark & \Checkmark & Euler angles & \color{blue}{\bf 2.876} & 4.924 & 3.303 & 3.701 & 3.473 & 4.817 & 2.769 & 3.686 \\
        \Xhline{1.2pt}
	\end{tabular}
	\end{center}
\end{table*}

\subsubsection{Single HPE in 300W-LP}
We compare our approach with the following three kinds of SOTA single HPE methods: {\it landmark-based}, {\it landmark-free} and {\it 3DMM-based}, which are introduced in Section \ref{HPE_DSL}. They may be trained only on the 300W-LP dataset, or utilize extra training data like WIDER FACE or self-built datasets. All results are collected in Table ~\ref{tabSingleHPE}. On the whole, we can summarize or verify three conclusions. (1) These landmark-based methods are the most vulnerable for their inherent disadvantages. For example, the top performer OsGG-Net \cite{mo2021osgg} is inferior to other methods in the same year. (2) All landmark-free methods with different head pose representations can achieve nearly equivalent performances. For example, without applying extra data, effects of MNN \cite{valle2020multi} using Euler angles can be approximate to vectors-based methods (Rankpose \cite{dai2020rankpose}, TriNet \cite{cao2021vector}, MFDNet \cite{liu2021mfdnet} and 6DRepNet \cite{hempel20226d}) and 3DMM-based methods (SynergyNet \cite{wu2021synergy} and DAD-3DNet \cite{martyniuk2022dad}). (3) Training with extra data and learning with multi-tasks are two pivotal factors for obtaining best results. For example, training with additionally self-labeled HPE dataset, 2DHeadPose \cite{wang20232dheadpose} achieves the second-best MAE {\bf 3.380} in AFLW2000 and the best MAE {\bf 3.440} in BIWI val-set. Comparatively, by combining multi-tasks including landmark detection, head pose estimation and 3D face reconstruction in a unified framework, SynergyNet \cite{wu2021synergy} obtains the best MAE {\bf 3.350} in AFLW2000 without extra dataset.

Compared with above SOTA methods, although not tailored for the single HPE task, our method directly trained on 300W-LP still achieves comparable MAE {\bf 4.035} in the val-set AFLW2000. Unsurprisingly, after pretraining on additional dataset WIDER FACE, our method can further reduce the MAE into {\bf 3.701}. It is better than the MAE {\bf 3.913} from the only MPHPE method Img2Pose \cite{albiero2021img2pose} which is also trained with extra WIDER FACE. We owe the superiority to our multi-task end-to-end pattern design.

\setlength{\tabcolsep}{1.1pt}
\begin{table}[!t]\scriptsize 
\begin{center}
\caption{Performance comparison on the validation set of AGORA-HPE benchmark. The ``---'' means not reported.}
\label{tabAGORA}
	\begin{tabular}{l|cccc|cc|rrrr}
	\Xhline{1.2pt}
	{\bf Method} & \makecell{{\bf Re-}\\{\bf train?}} & \makecell{{\bf Param}\\{\bf (MB)}} & \makecell{{\bf Input}\\ {\bf Size}} & \makecell{{\bf Latency}\\{\bf (ms)}}& {\bf AP} & {\bf $P_M$} & {\bf Pitch} & {\bf Yaw} & {\bf Roll} & {\bf MAE} \\
	\Xhline{1.2pt}
	\rowcolor{gray!15}
	\multicolumn{11}{c}{\bf  \textit{Narrow-range}: $-90^{\circ}<yaw<90^{\circ}$} \\
	\hline
	3DDFA \cite{zhu2016face} & No & 12.5 & 120 & 8.2 & --- & --- & 42.26 & 39.39 & 63.52 & 48.39 \\
	3DDFA\_v2 \cite{guo2020towards} & No & 13.1 & 120 & 3.2 & --- & --- & 20.53 & 27.82 & 18.98 & 22.44 \\
	SynergyNet \cite{wu2021synergy} & No & 71.9 & 120 & 7.2 & --- & --- & 35.58 & 39.55 & 51.51 & 42.21 \\
	DAD-3DNet \cite{martyniuk2022dad} & No & 126.0 & 256 & 18.4 & --- & --- & 38.89 & 19.99 & 39.04 & 32.64 \\
	\hline
	HopeNet \cite{ruiz2018fine} & No & 91.4 & 224 & 10.2 & --- & --- & 19.09 & 24.08 & 16.66 & 19.94 \\
	FSA-Net \cite{yang2019fsa} & No & 2.4 & 64 & 6.1 & --- & --- & 19.02 & 21.81 & 16.27 & 19.03 \\
	WHENet \cite{zhou2020whenet} & No & 16.5 & 224 & 50.4 & --- & --- & 20.92 & 35.30 & 15.79 & 24.00 \\
	Img2Pose \cite{albiero2021img2pose} & No & 161.0 & 400 & 18.1 & --- & --- & 22.15 & 17.44 & 20.24 & 19.94 \\
	6DRepNet \cite{hempel20226d} & No & 150.0 & 256 & 38.8 & --- & --- & 24.41 & 32.29 & 19.80 & 25.50 \\
	\hline
	FSA-Net \cite{yang2019fsa} & Yes & 2.9 & 64 & 4.6 & --- & --- & \color{red}{\bf 9.93} & \color{blue}{\bf 12.18} & \color{red}{\bf 9.55} & \color{blue}{\bf 10.55} \\
	6DRepNet \cite{hempel20226d} & Yes & 150.0 & 256 & 8.8 & --- & --- & \color{blue}{\bf 10.36} & 13.04 & \color{blue}{\bf 10.09} & 11.16 \\
	\hline
	DirectMHP-S & Yes & 25.3 & 1280 & 4.3 & 82.0 & 88.28 & 11.80 & 12.33 & 11.14 & 11.76  \\
	DirectMHP-M & Yes & 68.2 & 1280 & 9.6 & 83.5 & 90.32 & 11.19 & \color{red}{\bf 9.89} & 10.46 & \color{red}{\bf 10.51} \\
	\hline
	\rowcolor{gray!15}
	\multicolumn{11}{c}{\bf  \textit{Full-range}: $-180^{\circ}<yaw<180^{\circ}$} \\
	\hline
	WHENet \cite{zhou2020whenet} & No & 16.5 & 224 & 50.4 & --- & --- & 21.90 & 37.14 & 16.59 & 25.21 \\
	\hline
	6DRepNet \cite{hempel20226d} & Yes & 150.0 & 256 & 8.8 & --- & --- & \color{red}{\bf 11.82} & \color{blue}{\bf 14.91} & \color{blue}{\bf 11.08} & \color{blue}{\bf 12.60} \\
	\hline
	DirectMHP-S & Yes & 25.3 & 1280 & 4.3 & 74.5 & 87.20 & 12.98 & 14.98 & 11.72 & 13.23 \\
	DirectMHP-M & Yes& 68.2 & 1280 & 9.6 & 76.7 & 89.47 & \color{blue}{\bf 12.28} & \color{red}{\bf 12.42} & \color{red}{\bf 11.02} & \color{red}{\bf 11.94} \\
	\Xhline{1.2pt}
	\end{tabular}
\end{center}
\end{table}

\subsubsection{MPHPE in AGORA-HPE}
Table~\ref{tabAGORA} has shown the performance and latency of DirectMHP compared with other methods. Despite the need of detecting human heads and predicting head poses simultaneously, our one-stage approach still takes much less time than most methods. After reaching relatively high AP and $P_M$ values of head detection and matching, DirectMHP-M achieves the lowest MAE comparing with retained SOTA methods FSA-Net and 6DRepNet at narrow-range angles, and beats 6DRepNet at full-range angles. In general, four directly tested 3DMM-based methods achieve the worst results. They are usually sensitive to truncated, occluded, tiny or unevenly illuminated faces. However, we cannot retrain them for lacking 3DMM ground-truths of heads. Likewise, the five directly tested or two retrained landmark-free methods also perform worser than ours. This indicates that AGORA-HPE contains many challenging heads, which cannot be well tackled by previous single HPE approaches. For example, frustration is inevitable when predicting pose of well-cropped heads yet with heavy occlusion, low-resolution or bare facial information.

\setlength{\tabcolsep}{4pt}
\begin{table}[!t]\scriptsize 
\begin{center}
\caption{Performance comparison on the validation set of CMU-HPE benchmark. The ``---'' means not reported.}
\label{tabCMU}
	\begin{tabular}{l|c|cc|rrrr}
	\Xhline{1.2pt}
	{\bf Method} & {\bf Retrain?} & {\bf AP} & {\bf $P_M$} & {\bf Pitch} & {\bf Yaw} & {\bf Roll} & {\bf MAE} \\
	\Xhline{1.2pt}
	\rowcolor{gray!15}
	\multicolumn{8}{c}{\bf  \textit{Narrow-range}: $-90^{\circ}<yaw<90^{\circ}$} \\
	\hline
	3DDFA \cite{zhu2016face} & No & --- & --- & 26.11 & 23.43 & 31.24 & 26.93  \\
	3DDFA\_v2 \cite{guo2020towards} & No & --- & --- & 18.49 & 16.95 & 16.40 & 17.28 \\
	SynergyNet \cite{wu2021synergy} & No & --- & --- & 23.52 & 27.56 & 22.95 & 24.68 \\
	DAD-3DNet \cite{martyniuk2022dad} & No & --- & --- & 22.46 & 10.58 & 23.93 & 18.99 \\
	\hline
	HopeNet \cite{ruiz2018fine} & No & --- & --- & 17.31 & 20.05 & 13.33 & 16.90 \\
	FSA-Net \cite{yang2019fsa} & No & --- & --- & 16.28 & 17.49 & 12.91 & 15.56 \\
	WHENet \cite{zhou2020whenet} & No & --- & --- & 22.71 & 37.94 & 16.55 & 25.73 \\
	Img2Pose \cite{albiero2021img2pose} & No & --- & --- & 16.79 & 12.91 & 15.61 & 15.10 \\
	6DRepNet \cite{hempel20226d} & No & --- & --- & 15.21 & 16.65 & 13.53 & 15.13 \\
	\hline
	FSA-Net \cite{yang2019fsa} & Yes & --- & --- & 9.74 & 8.31 & 7.92 & 8.66 \\
	6DRepNet \cite{hempel20226d} & Yes & --- & --- & \color{red}{\bf 7.21} & \color{red}{\bf 4.98} & \color{red}{\bf 6.06} & \color{red}{\bf 6.08} \\
 	\hline
	DirectMHP-S & Yes & 84.3 & 96.07 & \color{blue}{\bf 8.01} & \color{blue}{\bf 5.75} & \color{blue}{\bf 6.96} & \color{blue}{\bf 6.91}  \\
	DirectMHP-M & Yes & 85.8 & 96.29 & 8.13 & 6.02 & 7.08 & 7.08 \\
	\Xhline{1.2pt}
	\rowcolor{gray!15}
	\multicolumn{8}{c}{\bf  \textit{Full-range}: $-180^{\circ}<yaw<180^{\circ}$} \\
	\hline
	WHENet \cite{zhou2020whenet} & No & --- & --- & 19.89 & 29.89 & 14.67 & 21.48 \\
	\hline
	6DRepNet \cite{hempel20226d} & Yes & --- & --- & \color{red}{\bf 7.73} & \color{red}{\bf 5.73} & \color{red}{\bf 6.51} & \color{red}{\bf 6.66} \\
	\hline
	DirectMHP-S & Yes & 80.8 & 97.39 & \color{blue}{\bf 8.54} & 7.32 & \color{blue}{\bf 7.35} & \color{blue}{\bf 7.74} \\
	DirectMHP-M & Yes & 82.5 & 97.67 & 8.64 & \color{blue}{\bf 7.14} & 7.51 & 7.76 \\
	\Xhline{1.2pt}
	\end{tabular}
\end{center}
\end{table}

\subsubsection{MPHPE in CMU-HPE}
The performance of DirectMHP and compared methods on CMU-HPE is shown in Table~\ref{tabCMU}. Similar to AGORA-HPE, our method is still very superior in every metric. The difference is that the DirectMHP-S with less parameters achieves lower MAE values than DirectMHP-M. Actually, more detected and matched heads containing hard samples by DirectMHP-M ordinarily lead to a larger MAE. Another anomaly is that 6DRepNet performs better than our method. We assume that this is because CMU-HPE lacks the diversity of background and subjects, resulting in over-fitting of 6DRepNet. On the other hand, the weakness of our Euler angles based method occurs when the yaw is close to $\pm90^{\circ}$. Furthermore, due to the simplicity of CMU-HPE, we consistently obtain lower MAE values than in AGORA-HPE. More details are analyzed in Section \ref{sectionEA}.

\begin{figure*}[t]
	\centering
	\subfloat[]{\includegraphics[width=0.325\textwidth]{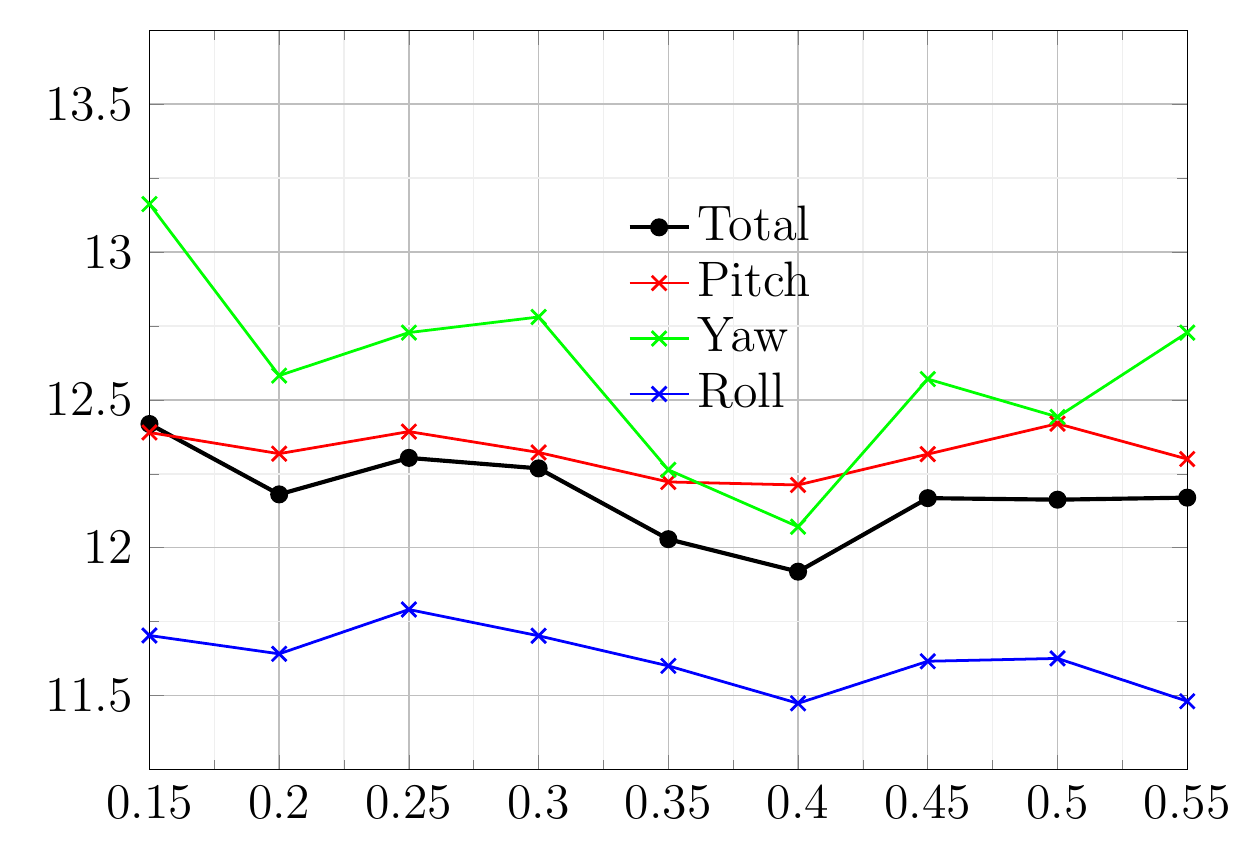}
	\label{tauAS}}
	\hfill
	\subfloat[]{\includegraphics[width=0.325\textwidth]{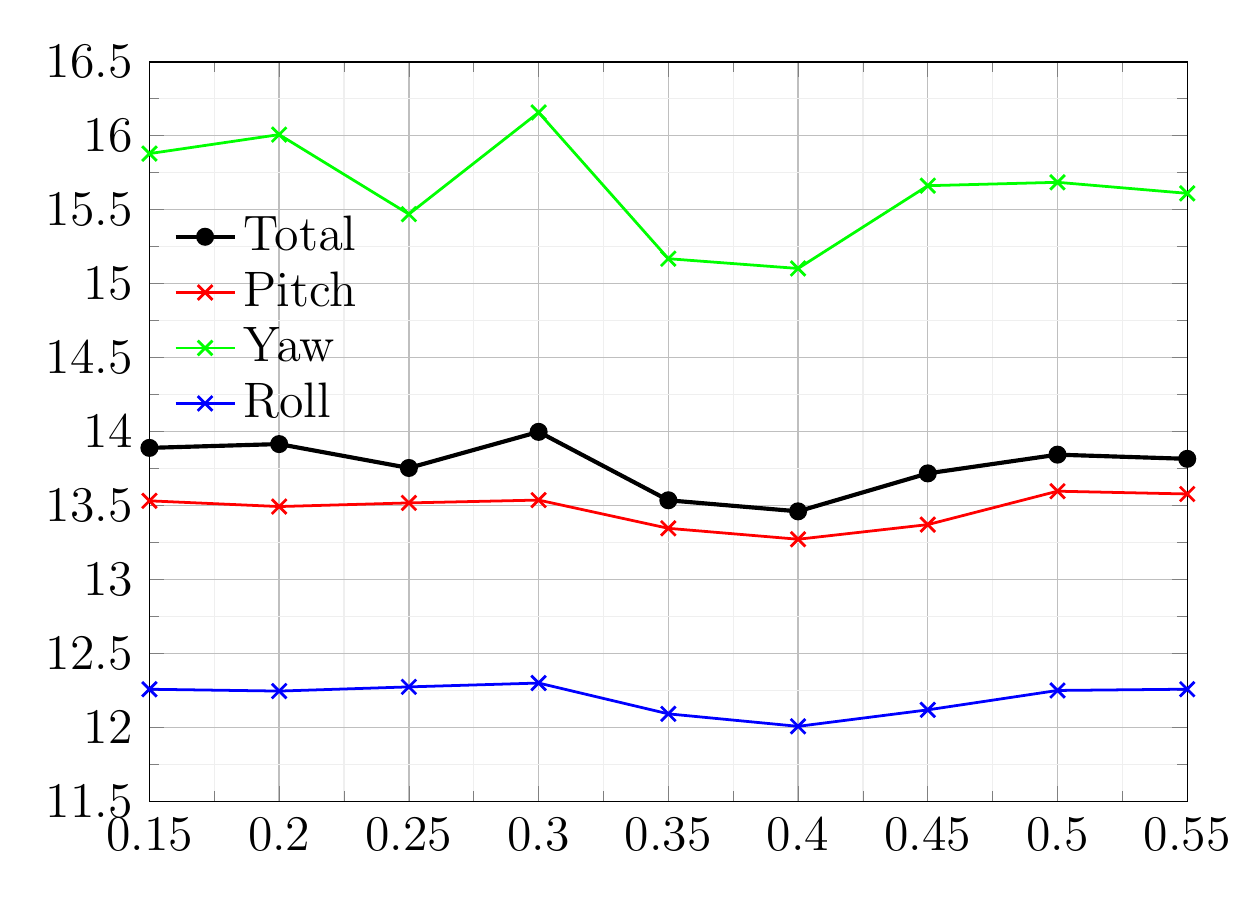}
	\label{tau2AS}}
	\hfill
	\subfloat[]{\includegraphics[width=0.325\textwidth]{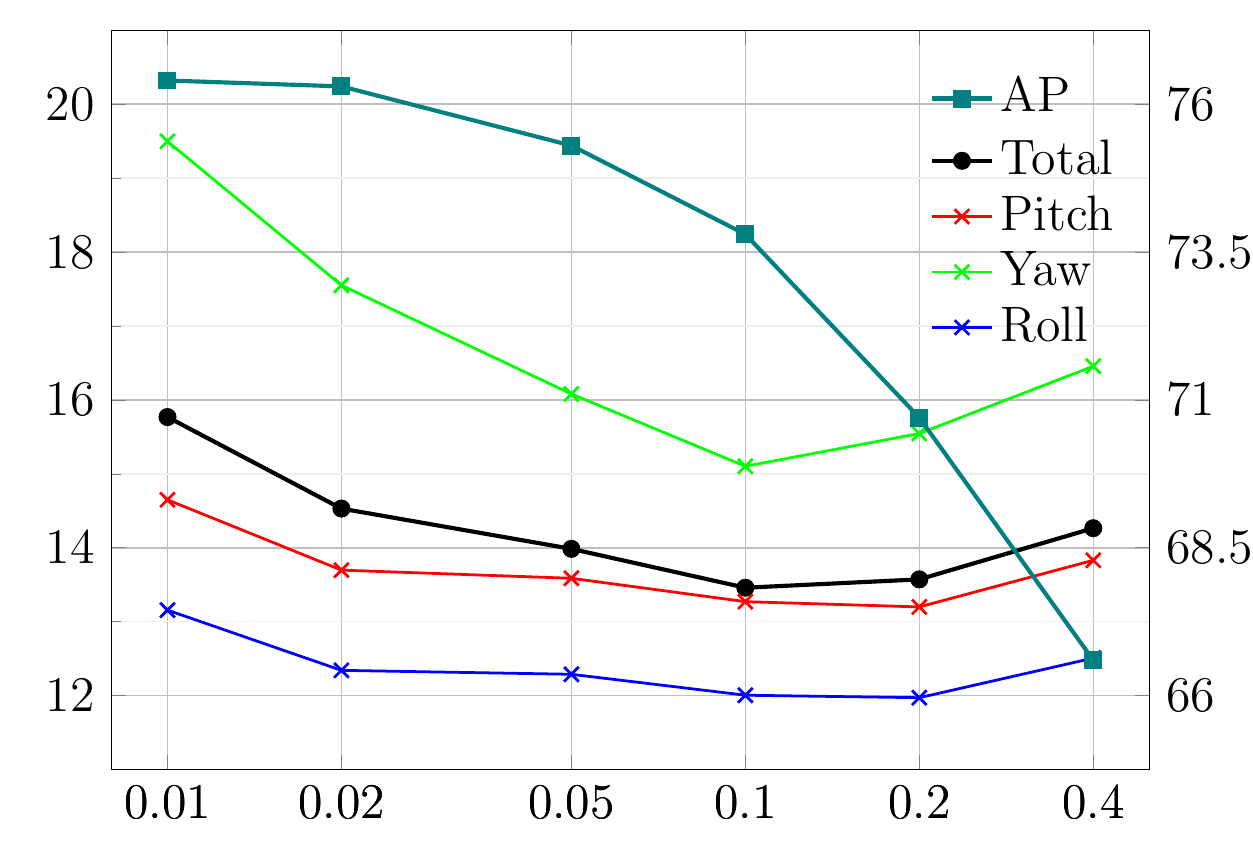}
	\label{gammaAS}}
	\caption{{\bf (a)} The influence of $\tau$ (x-axis) on {\em narrow-range} MAE results (y-axis). {\bf (b)} The influence of $\tau$ (x-axis) on {\em full-range} MAE results (y-axis). {\bf (c)} The influence of loss weight $\gamma$ (x-axis) on full-range MAE results (left y-axis) and AP (right y-axis).}
	\label{AS}
\end{figure*}

\subsection{Ablation Studies}\label{sectionAS}

We mainly investigate three parameters: tolerance threshold $\tau$, loss weight $\gamma$, and whether using the wrapped MSE loss. If not specifically emphasized, we conduct all ablation experiments on the AGORA-HPE dataset using DirectMHP-S. The total epoch is reduced to 100 for simplicity.

\subsubsection{Tolerance Threshold $\tau$}
The parameter $\tau$ is used to select high-scoring candidates in loss $\mathcal{L}_{mse}$. As shown in Fig.~\ref{tauAS}, we uniformly sample $\tau$ from 0.15 to 0.55 for training, and report the narrow-range MAE results of each last model. By primitively setting $\gamma=0.1$, we can clearly find that the model performance is optimal when $\tau=0.4$. A larger or smaller $\tau$ will lead to inferior results. The same conclusion is true when we observe the variation of full-range MAE results as plotted in Fig.~\ref{tau2AS}. This reminds us to choose an appropriate $\tau$ to ensure that enough head object samples are left to facilitate steady convergence of training, while removing redundant destructive head proposals.

\subsubsection{Loss Weight $\gamma$}
For the weight $\gamma$ in the total loss $\mathcal{L}$, it should keep a balance with $\alpha$ and $\beta$ which have been fine-tuned and predetermined in YOLOv5. As shown in Fig.~\ref{gammaAS}, we set $\tau=0.4$ and select $\gamma$ among six proportionally increasing values. We additionally add an update to the metric AP. Finally, we achieve a compromise balance between MAE and AP when $\gamma$ is properly set to 0.1.

\subsubsection{Wrapped MSE Loss}
By fixing $\tau$ and $\gamma$, we separately choose to use the MSE loss $\mathcal{L}_{mse}$ or wrapped MSE loss $\mathcal{L}_{wmse}$ when training. We show the comparison of all metrics under three control groups in Table~\ref{tab3}. Basically, after using the wrapped loss, the performance of each metric in all groups is always worse (including higher MAE and lower AP values). Therefore, we drop $\mathcal{L}_{wmse}$ in training and only consider wrapped angles when measuring MAE.

\setlength{\tabcolsep}{8pt}
\begin{table}[!t]\scriptsize 
\begin{center}
\caption{The influence of wrapped MSE (WMSE) loss on AP and full-range MAE results.}
\label{tab3}
	\begin{tabular}{ccc|c|cccc}
	\Xhline{1.2pt}
	{\bf WMSE} & $\tau$ & $\gamma$ & {\bf AP} & {\bf Pitch} & {\bf Yaw} & {\bf Roll} & {\bf MAE} \\
	\Xhline{1.2pt}
	Yes & 0.40 & 0.10 & 72.1 & 13.97 & 16.38 & 12.60 & 14.32  \\
	No & 0.40 & 0.10 & {\bf 73.8} & 13.27 & 15.10 & 12.01 & {\bf 13.46} \\
	\hline
	Yes & 0.35 & 0.10 & 70.9 & 14.00 & 16.69 & 12.65 & 14.45  \\
	No & 0.35 & 0.10 & {\bf 73.4} & 13.35 & 15.17 & 12.09 & {\bf 13.54} \\
	\hline
	Yes & 0.40 & 0.05 & 74.6 & 13.74 & 16.40 & 12.44 & 14.19  \\
	No & 0.40 & 0.05 & {\bf 75.3} & 13.59 & 16.08 & 12.29 & {\bf 13.99} \\
	\Xhline{1.2pt}
	\end{tabular}
\end{center}
\end{table}

\subsection{Error Analysis}\label{sectionEA}

\subsubsection{AGORA-HPE}
\begin{figure}[!]
\centering
\includegraphics[width=0.9\columnwidth]{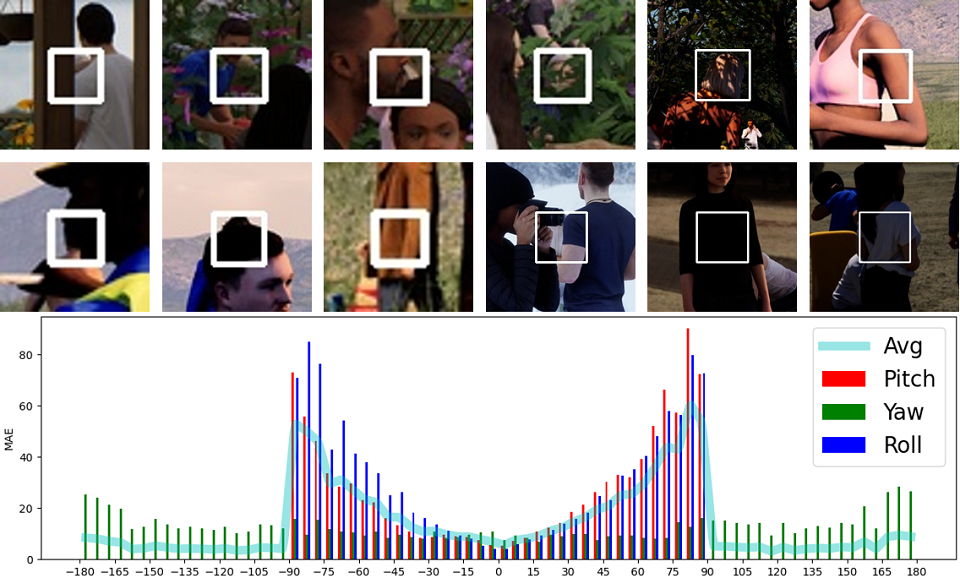}
\caption{{\bf Top:} Examples of undetected human heads on the val-set of AGORA-HPE. {\bf Bottom:} Histogram of the MAE on the val-set of AGORA-HPE.}
\label{AGORAUNDetErr}
\end{figure}

For our best performed full-range model DirectMHP-M, we have totally detected 6,715 heads out of 7,505 labeled heads in the val-set of AGORA-HPE ($P_M$=89.47\%). We picked out some typical examples from those undetected heads as shown in the top of Fig.~\ref{AGORAUNDetErr}. In summary, these hard examples contain a variety of {\em severely occluded}, {\em low-resolution}, or {\em background-confused} heads, which CNN-based object detection methods have not been able to address well so far. For these detected heads, we plots the MAE of predicted three Euler angles with ground-truths at different degree intervals as shown in the bottom of Fig.~\ref{AGORAUNDetErr}. The MAE of yaw angle is basically consistent across full-range views. However, extremely high MAE of pitch or roll occurs when the yaw angle approaches $\pm90^{\circ}$. This is caused by the {\em gimbal lock} problem, which is an inherent flaw in the representation of 2D head pose estimation. 

\subsubsection{CMU-HPE}
\begin{figure}[!]
\centering
\includegraphics[width=0.9\columnwidth]{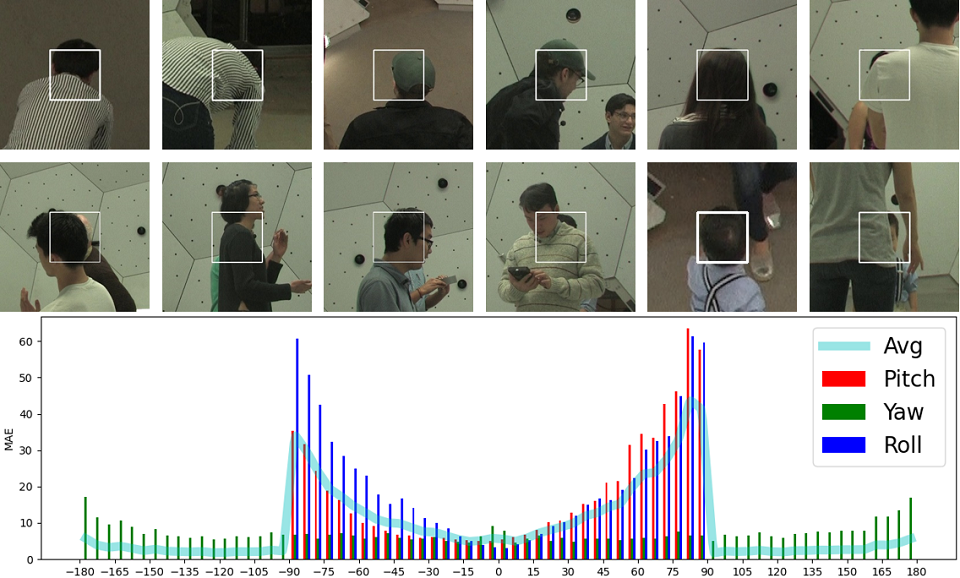}
\caption{{\bf Top:} Examples of undetected human heads on the val-set of CMU-HPE. {\bf Bottom:} Histogram of the MAE on the val-set of CMU-HPE.}
\label{CMUUNDetErr}
\end{figure}

Similarly, we have totally detected 31,976 heads out of 32,738 heads in the validation set of CMU-HPE ($P_M$=97.67\%) by using the best model DirectMHP-M. Undetected head examples are shown in the top of Fig.~\ref{CMUUNDetErr}. These difficult samples are mainly affected by factors including {\em severe occlusion}, {\em extreme views}, and {\em inconspicuous human head features}. For these detected heads, the MAE distributions of their estimated poses are shown in the bottom of Fig.~\ref{CMUUNDetErr}. We can observe similar conclusions to distributions on AGORA-HPE. The difference is that the MAE values on the simpler CMU-HPE dataset are always on the small side.

\subsubsection{Comparing with 6DRepNet}
\begin{figure}[!]
\centering
\includegraphics[width=0.9\columnwidth]{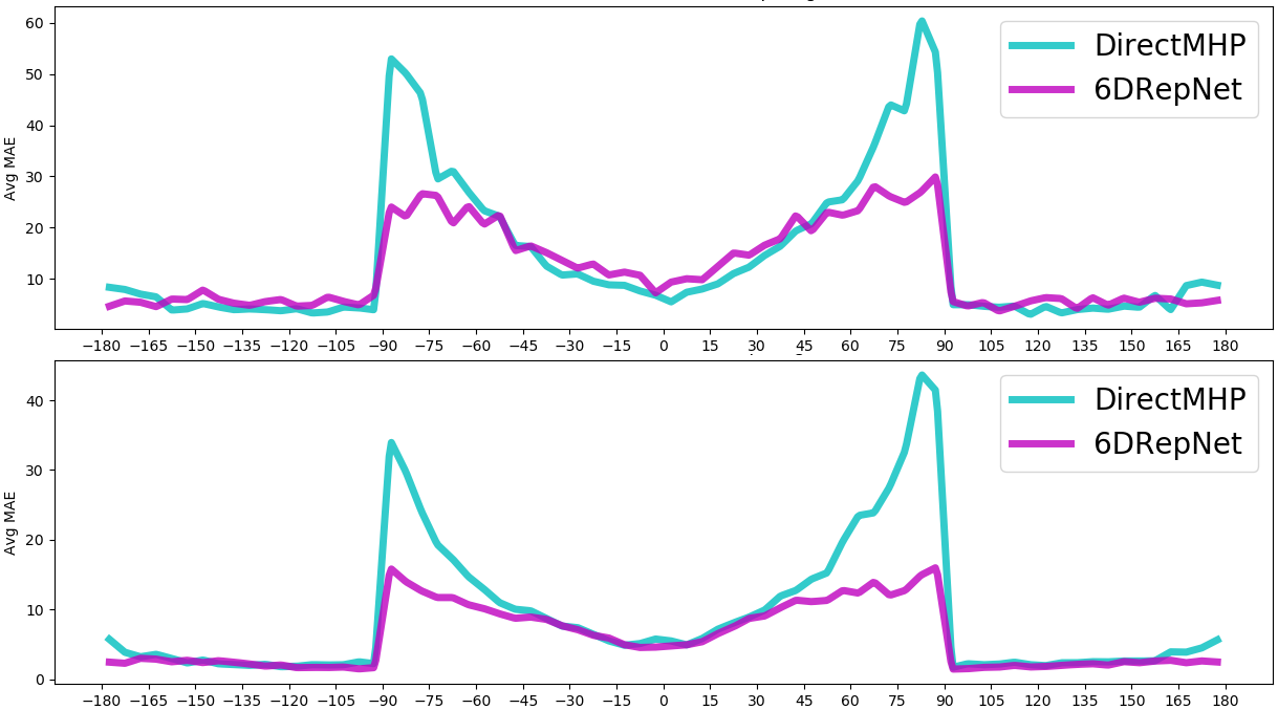}
\caption{{\bf Top:} The MAE comparing on the val-set of AGORA-HPE. {\bf Bottom:} The MAE comparing on the val-set of CMU-HPE.}
\label{ErrComparing}
\end{figure}

In order to understand the difference between the SOTA method 6DRepNet and our DirectMHP in obtaining comparable MAE results, we collected their error distributions on the two val-sets AGORA-HPE and CMU-HPE. As shown in Fig.~\ref{ErrComparing}, 6DRepNet performed much better than our method when the yaw angle approaches $\pm90^{\circ}$. This is due to its superior vectors representation. However, in the more challenging AGORA-HPE val-set, even with the inferior Euler angle representation, our method obtained lower MAE when the yaw angle approaches $0^{\circ}$ or $\pm180^{\circ}$.

\subsection{Qualitative Comparison}

\begin{figure}[!]
\includegraphics[width = 0.495\columnwidth]{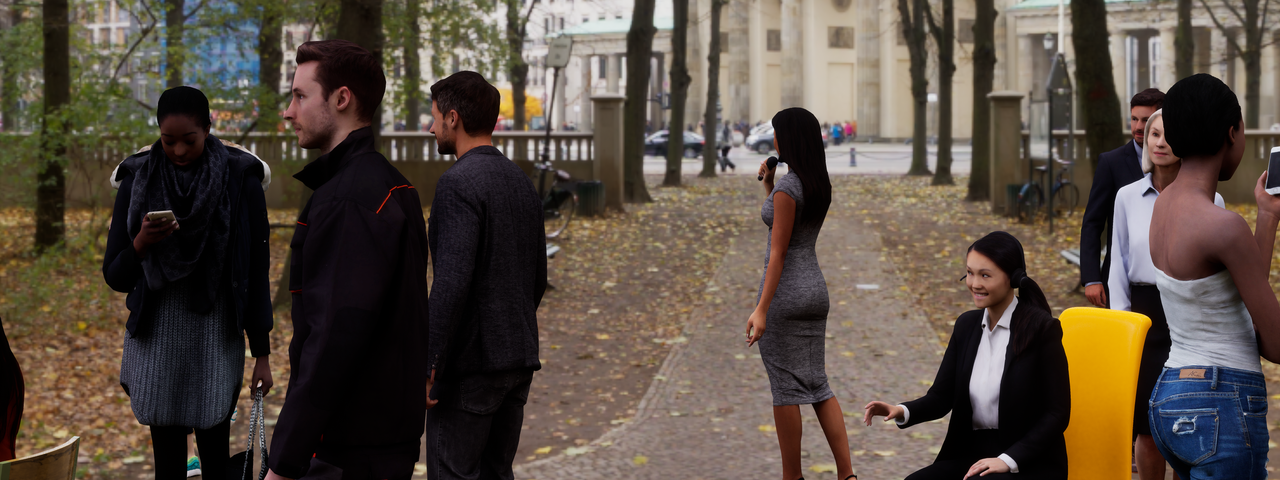}
\includegraphics[width = 0.495\columnwidth]{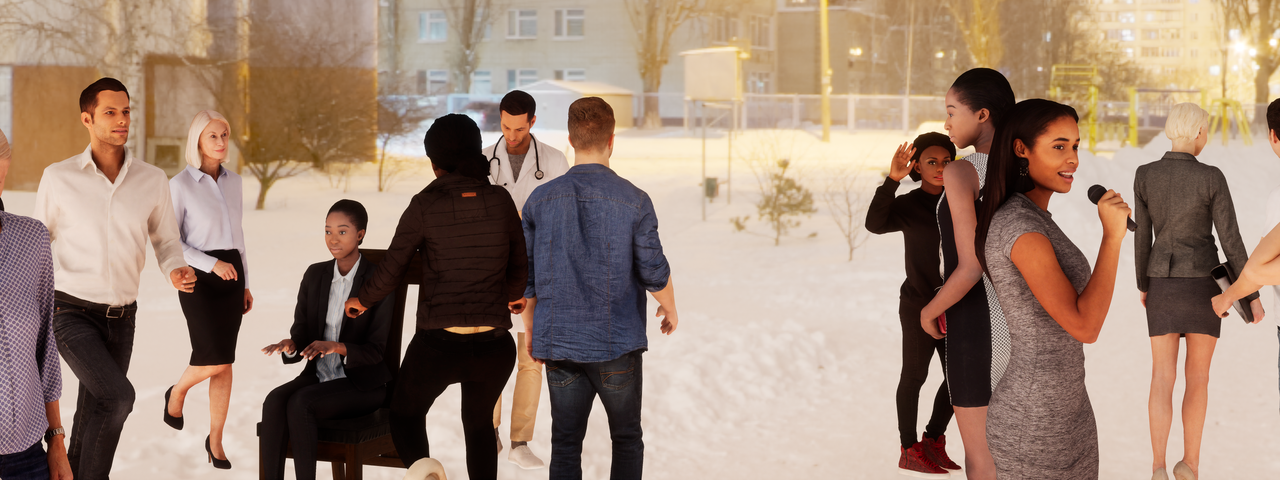}\smallskip \\
\includegraphics[width = 0.495\columnwidth]{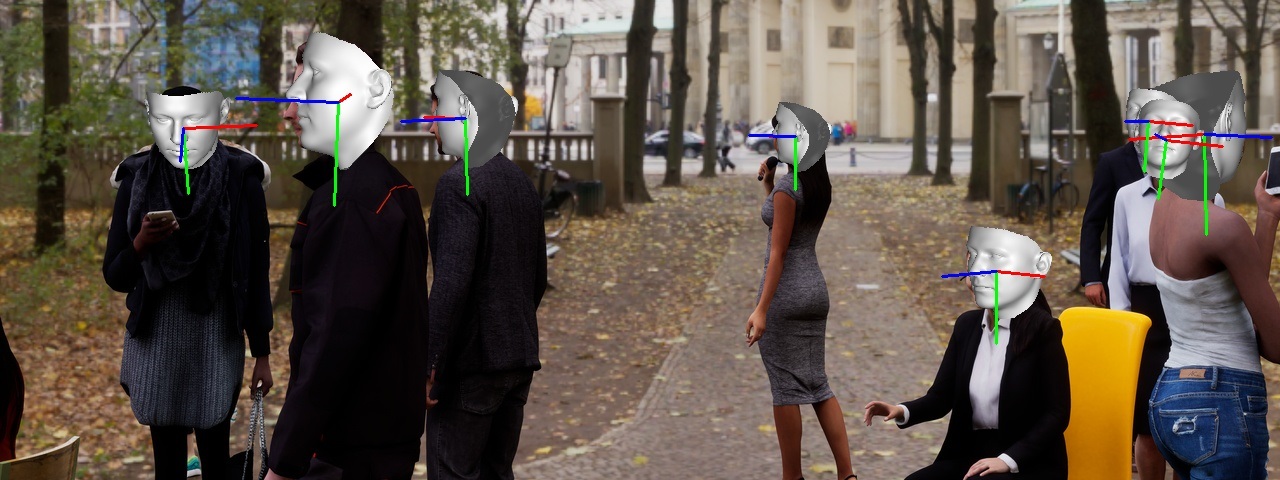}
\includegraphics[width = 0.495\columnwidth]{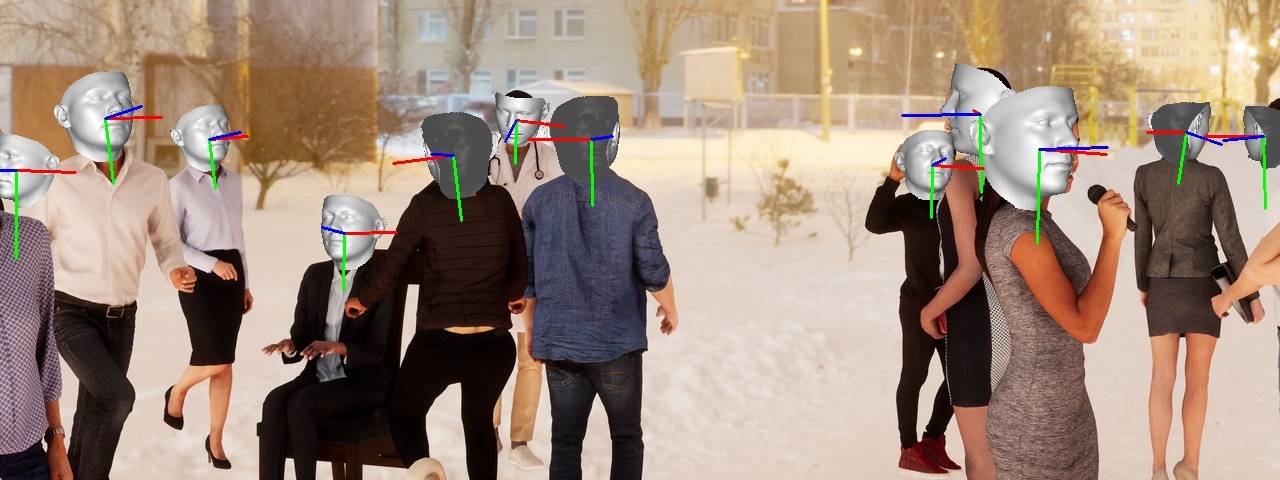}\medskip \\
\includegraphics[width = 0.327\columnwidth]{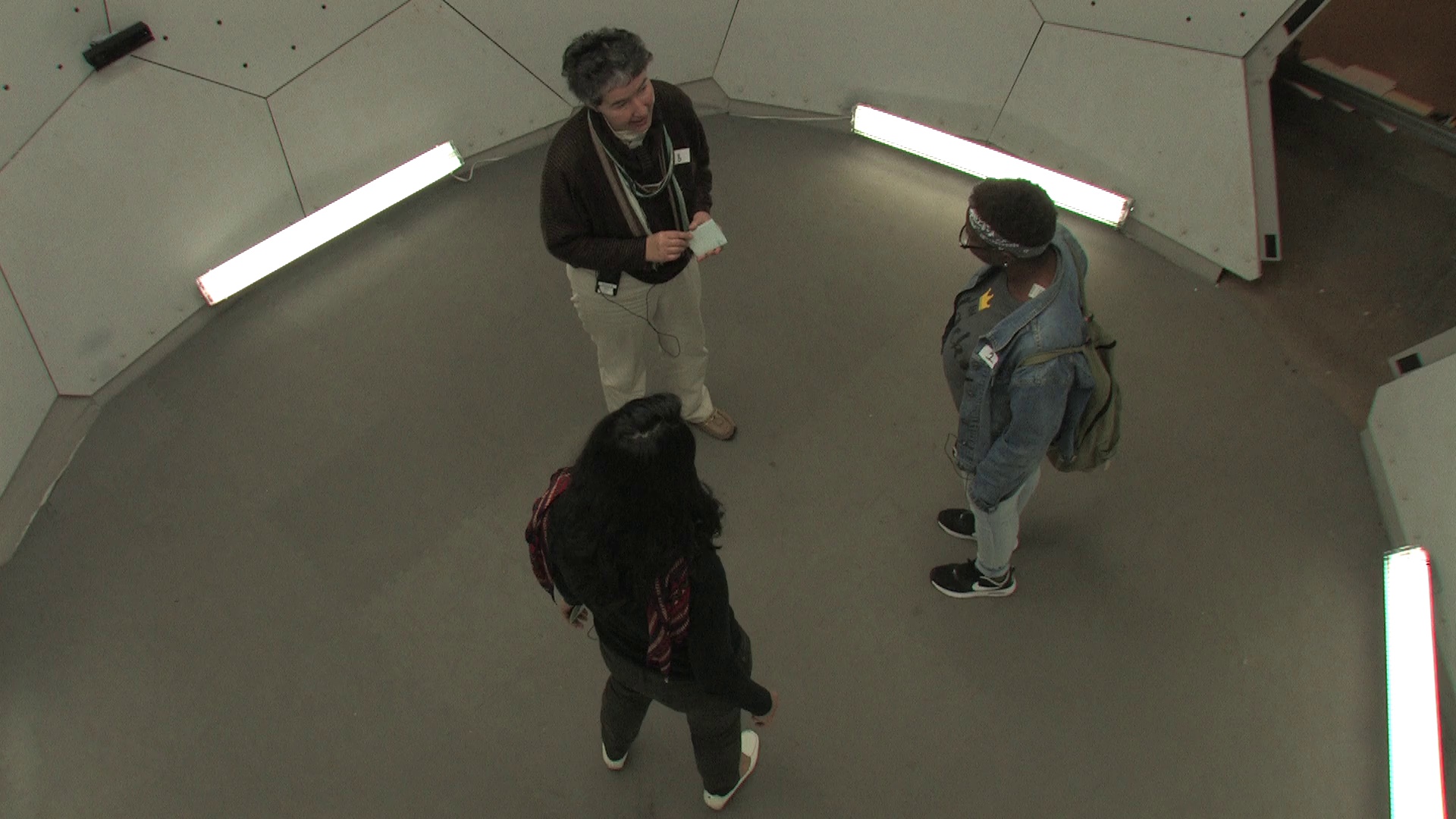}
\includegraphics[width = 0.327\columnwidth]{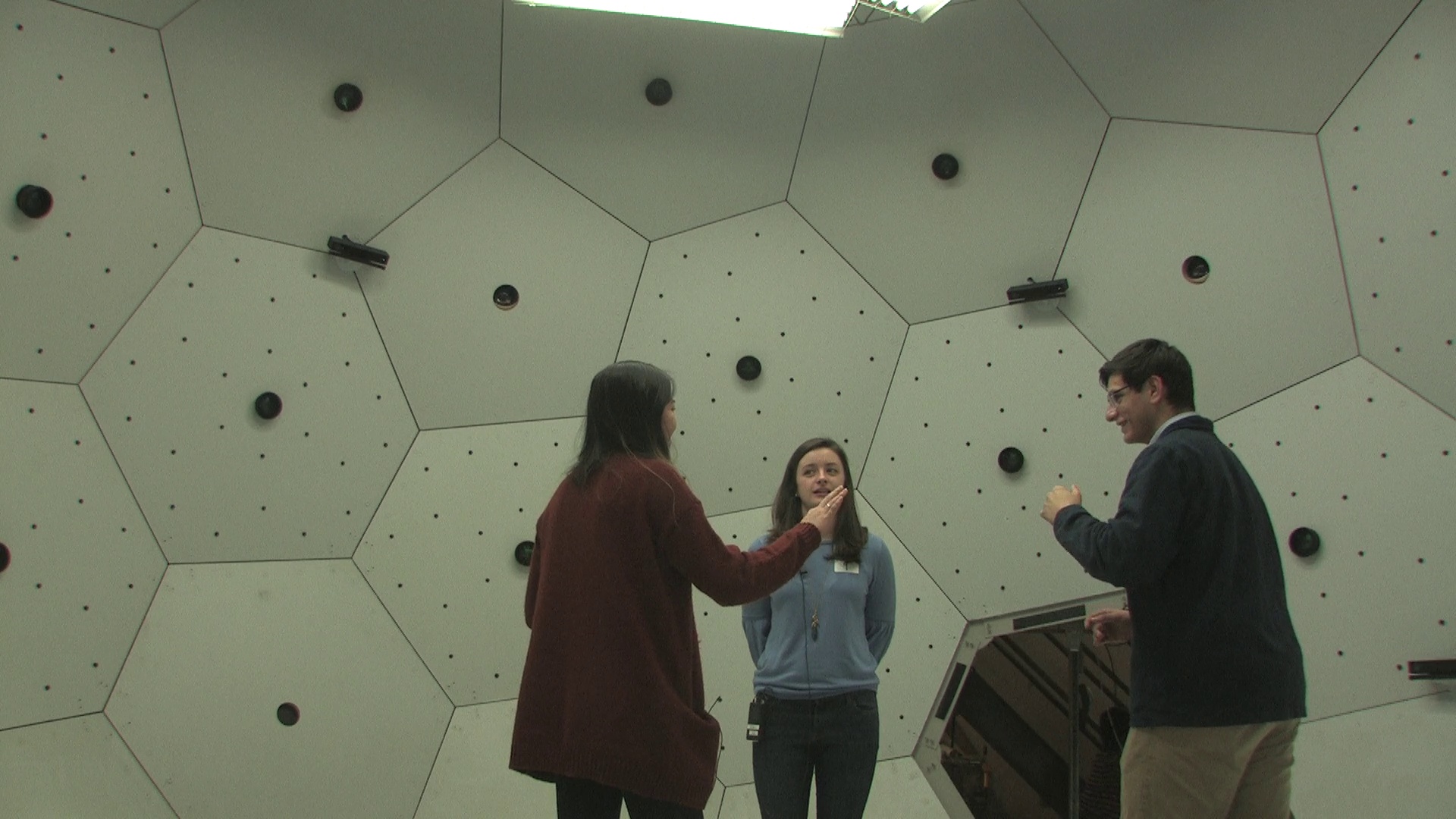}
\includegraphics[width = 0.327\columnwidth]{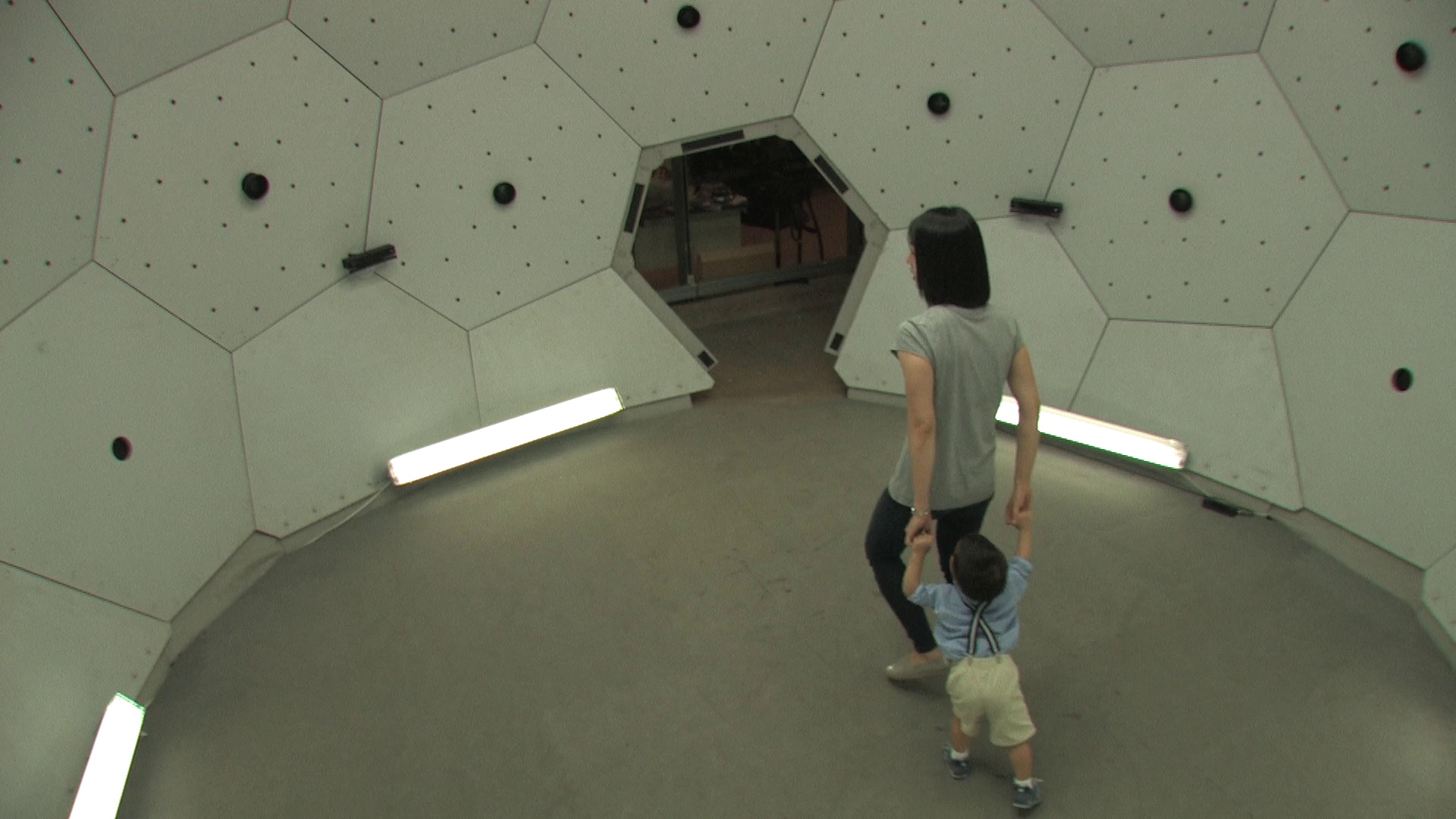}\smallskip \\
\includegraphics[width = 0.327\columnwidth]{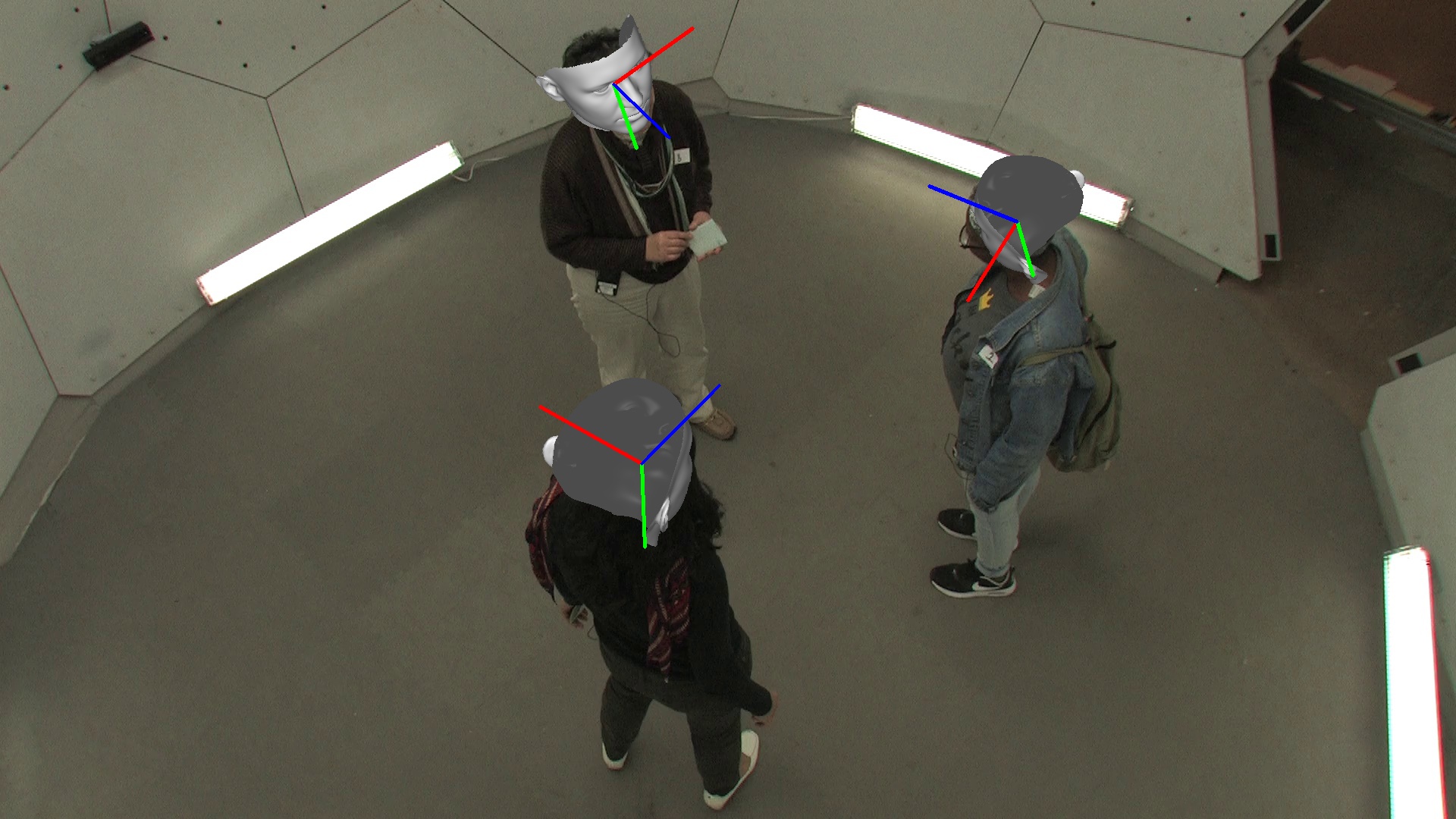}
\includegraphics[width = 0.327\columnwidth]{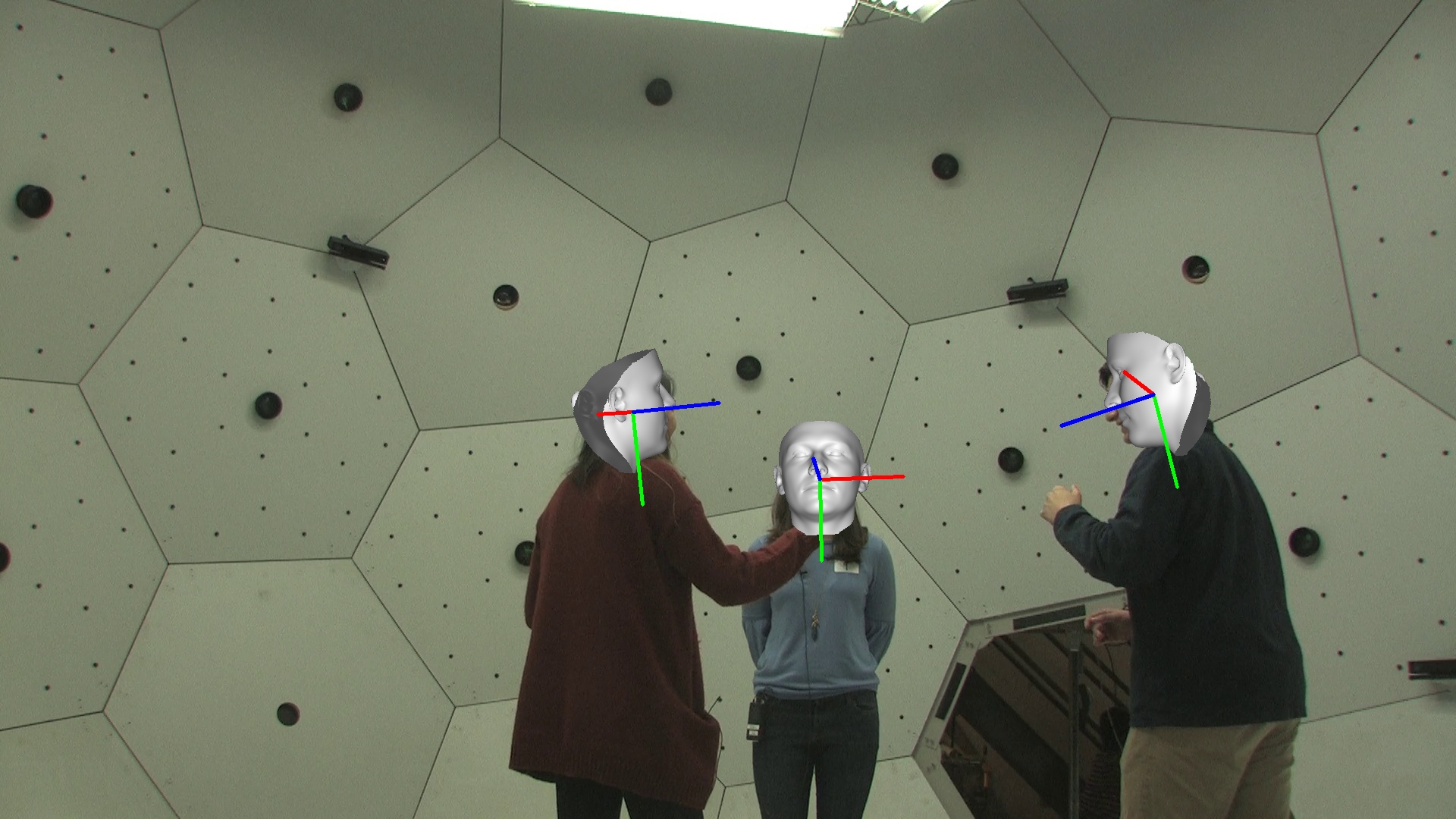}
\includegraphics[width = 0.327\columnwidth]{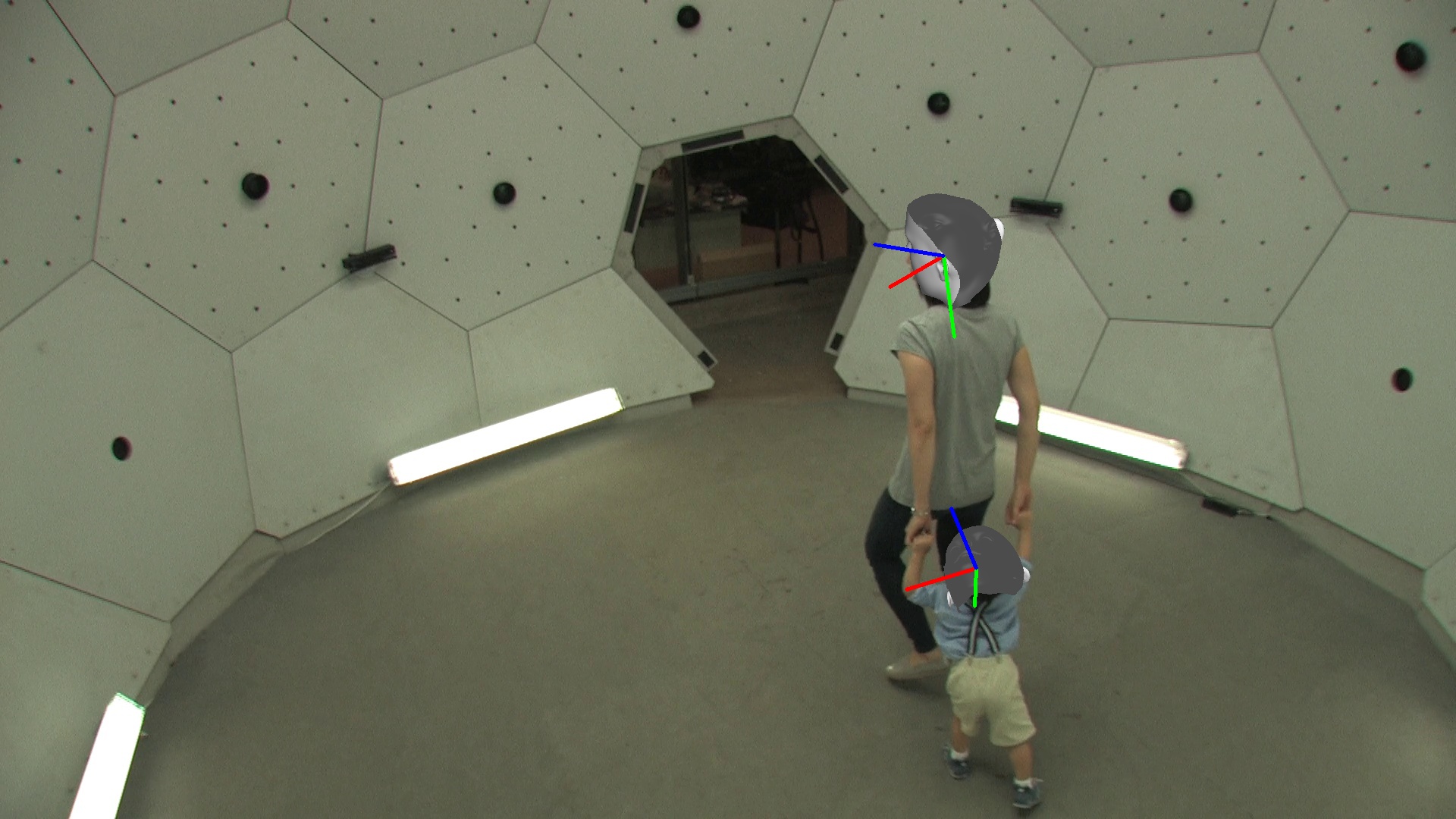}
\caption{Visualization on AGORA-HPE ({\it second row}) and CMU-HPE ({\it fourth row}) validation images. Head pose label is plotted by pitch (red-axis), yaw (green-axis) and roll (blue-axis) angles in indicated directions. These projected 2D face meshes are rendered by the Sim3DR tools as in 3DDFA\_v2 \cite{guo2020towards}. Note that these 2D meshes are only for more intuitive visualization of head pose, not for face reconstruction or dense face alignment tasks.}
\label{visualizationA}
\end{figure}

In Fig.~\ref{visualizationA}, we show representative head detection and pose estimation results on AGORA-HPE and CMU-HPE validation images produced by our model DirectMHP-M. After training and testing on data with the same domain distribution, our model has achieved satisfactory prediction accuracy. Meanwhile, it has the advantage of efficient end-to-end simultaneous head pose estimation of multi-person.

\begin{figure*}[!]
\includegraphics[height = 0.248\columnwidth]{figures/COCO/000000002685}
\includegraphics[height = 0.248\columnwidth]{figures/COCO/000000018380}
\includegraphics[height = 0.248\columnwidth]{figures/COCO/000000038829}
\includegraphics[height = 0.248\columnwidth]{figures/COCO/000000081988}
\includegraphics[height = 0.248\columnwidth]{figures/COCO/000000161925}
\includegraphics[height = 0.248\columnwidth]{figures/COCO/000000183648}\smallskip\\ 
\includegraphics[height = 0.248\columnwidth]{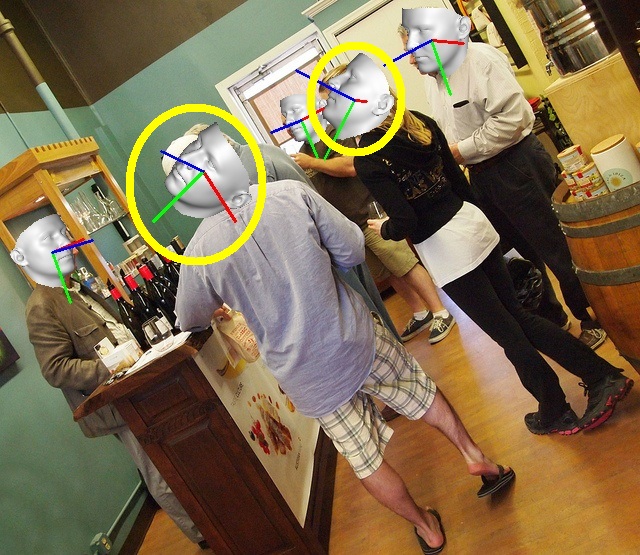}
\includegraphics[height = 0.248\columnwidth]{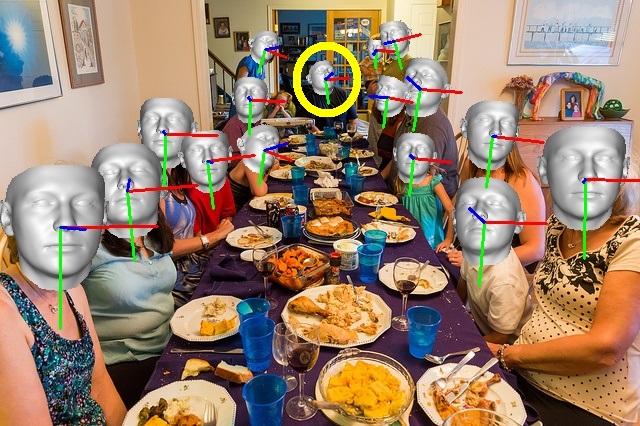}
\includegraphics[height = 0.248\columnwidth]{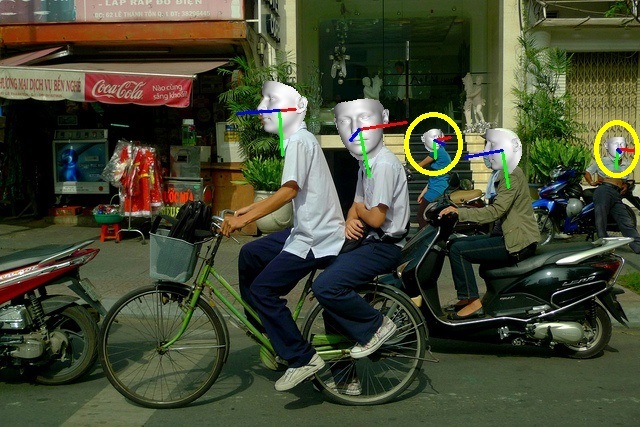}
\includegraphics[height = 0.248\columnwidth]{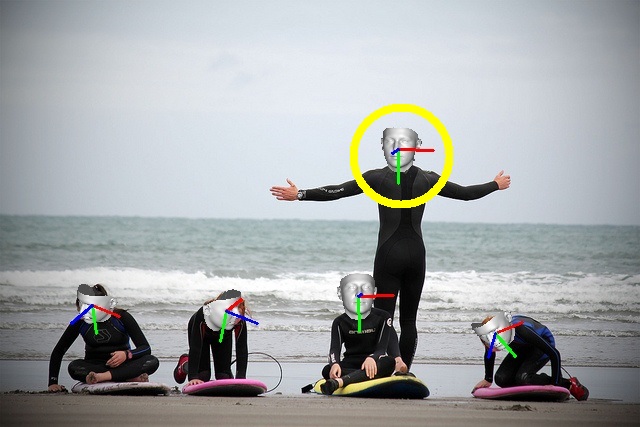}
\includegraphics[height = 0.248\columnwidth]{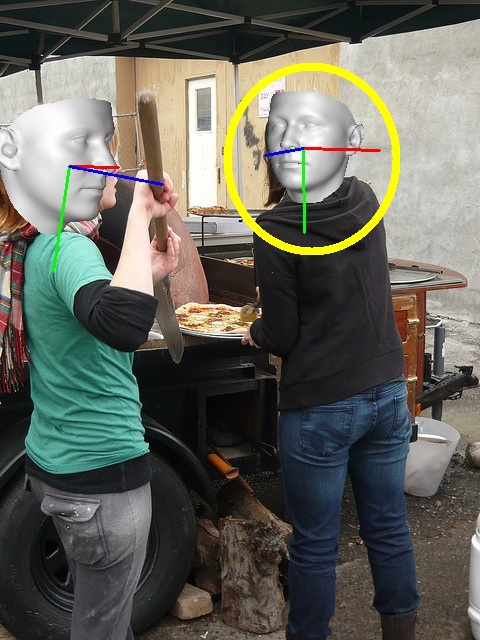}
\includegraphics[height = 0.248\columnwidth]{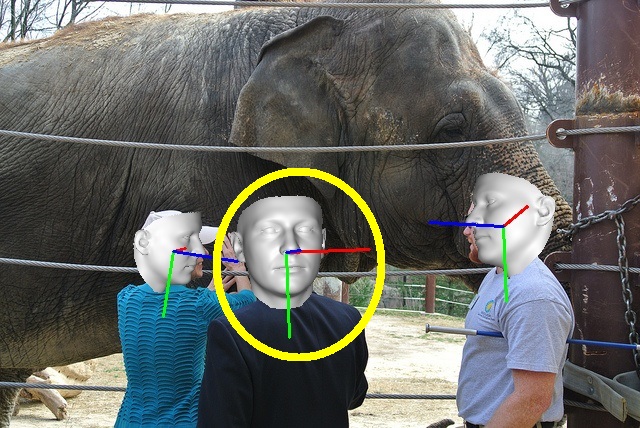}\smallskip\\
\includegraphics[height = 0.248\columnwidth]{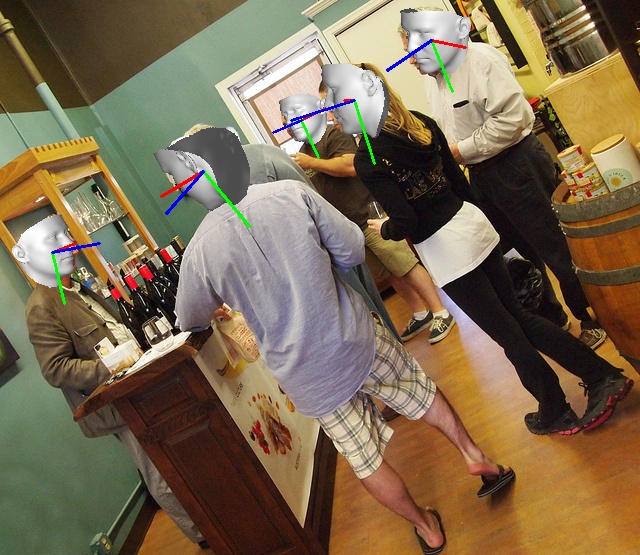}
\includegraphics[height = 0.248\columnwidth]{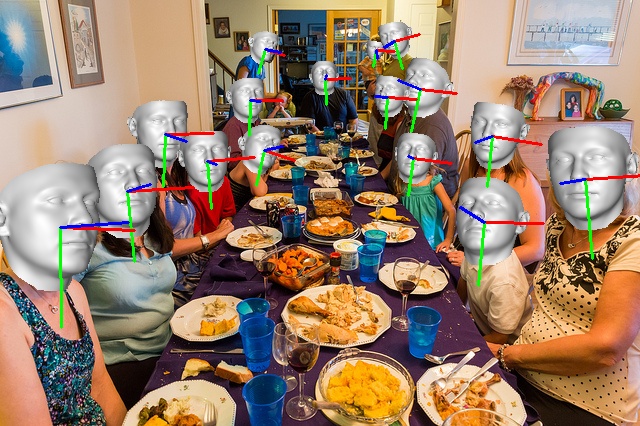}
\includegraphics[height = 0.248\columnwidth]{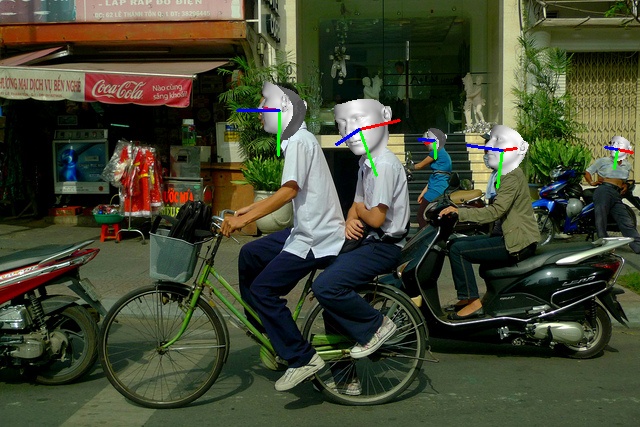}
\includegraphics[height = 0.248\columnwidth]{figures/COCO/000000081988_vis3d_res}
\includegraphics[height = 0.248\columnwidth]{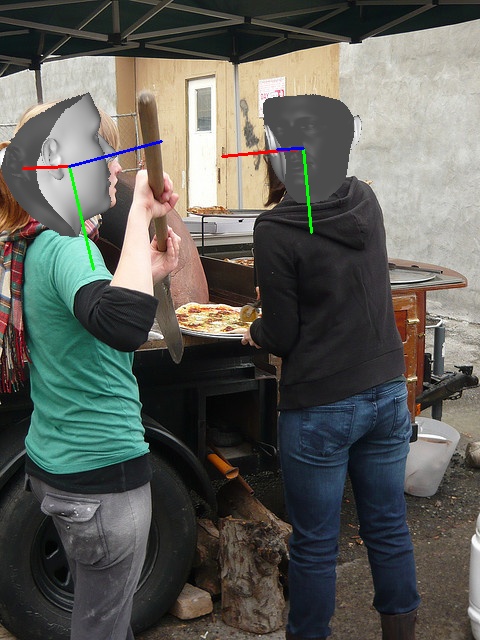}
\includegraphics[height = 0.248\columnwidth]{figures/COCO/000000183648_vis3d_res}
\caption{Visualization on some in-the-wild images from COCO {\it val-set}. The {\it second} and {\it third} rows are examples by compared 6DRepNet and our method, respectively. For those estimated head samples using 6DRepNet with obvious inaccuracies, we circled them in yellow color. Since 6DRepNet is a single-person HPE method, we appropriately cropped each detected head by our DirectMHP-M model as its input.}
\label{visualizationB}
\end{figure*}

Furthermore, to emphasize the generality and robustness of our method in unseen domains, we compare our DirectMHP-M with the full-range 6DRepNet \cite{hempel20226d} visually on some images selected from the COCO \cite{lin2014microsoft} {\it val-set} that contain multiple people but no head pose labels. Both models are trained on AGORA-HPE. As shown in Fig.~\ref{visualizationB}, in some challenging cases, especially the head sample with invisible face ({\it columns 1, 4, 5 and 6}), our method generally does not have significant head orientation angle errors like 6DRepNet. This is largely supported by our end-to-end design which can exploit the relation of each head with its whole human body in the original images. For these heads with frontal faces, 6DRepNet can obtain more refined head poses in some ordinary cases than our method, which corresponds to its competitive MAE results reported in Table~\ref{tabAGORA}. However, when encountering rare affine deformation ({\it column 1}) or low resolution ({\it columns 2 and 3}), our method will be more robust and reliable.

\section{Discussion and Conclusions}
We initially point out the MPHPE problem, and attempt to address it by training the proposed DirectMHP model on two reconstructed benchmarks. However, both datasets presented in this paper have limitations. The AGORA-HPE is a synthetic dataset using finite virtual 3D human models and backgrounds. Images in CMU-HPE are collected in an invariant controlled dome where sparse persons perform non-spontaneous behaviors. They are not the ultimate substitute for datasets with complex wild environments.

We envision that these problems can be alleviated in the future by constructing real-world MPHPE datasets. For example, DAD-3DHeads \cite{martyniuk2022dad} is a recently released 3D head dataset with manually annotated 3D head meshes labels in 2D in-the-wild images. 2DHeadPose \cite{wang20232dheadpose} has built a dataset by manually simulating the head pose in 2D in-the-wild images using a 3D virtual human head. However, both of them are limited to the single-person HPE task instead of MPHPE. Besides, considering the labor cost and ambiguity of directly labeling 2D head poses, we may turn to semi-supervised learning or domain adaptation methods which can excavate the potential of large amounts of unlabeled 2D images.

In summary, we propose a novel end-to-end network structure for tackling the multi-person head pose estimation (MPHPE) problem under full-range angles. By implicitly utilizing context information around person head, our one-stage baseline can achieve impressive results on two reconstructed dedicated datasets. Comprehensive ablation studies further confirm the robustness and reliability of our method. We expect our work to contribute to a better development in the field of MPHPE.


\section*{Acknowledgments}
This paper is supported by NSFC (No. 62176155), Shanghai Municipal Science and Technology Major Project, China, under grant no. 2021SHZDZX0102. 

\bibliographystyle{IEEEtran}
\bibliography{refs}

\end{document}